\documentclass[10pt,twocolumn,letterpaper]{article}

\usepackage{wacv}              

\usepackage{graphicx}
\usepackage{amsmath}
\usepackage{amssymb}
\usepackage{booktabs}

\usepackage{algorithm}
\usepackage{algpseudocode}
\usepackage{mathtools}

\usepackage[pagebackref,breaklinks,colorlinks]{hyperref}

\usepackage[capitalize]{cleveref}
\crefname{section}{Sec.}{Secs.}
\Crefname{section}{Section}{Sections}
\Crefname{table}{Table}{Tables}
\crefname{table}{Tab.}{Tabs.}

\def\wacvPaperID{474} 
\def\confName{WACV}
\def\confYear{2025}

\newcommand{\ourname}{FitDiff}

\algnewcommand\algorithmicforeach{\textbf{for all}}
\algdef{S}[FOR]{ForEach}[1]{\algorithmicforeach\ #1\ \algorithmicdo}

\begin{document}

\title{\ourname{}: Robust monocular 3D facial shape \\
 and reflectance estimation using Diffusion Models}


\author{Stathis Galanakis$^{1,2}$
\and
Alexandros Lattas$^{1}$
\and
Stylianos Moschoglou$^{1}$
\and
Stefanos Zafeiriou$^{1}$
\and
\\
$^1$Imperial College London \\
$^2$HUAWEI Noah's Ark Lab 
}

\maketitle

\begin{abstract}
The remarkable progress in 3D face reconstruction has resulted in high-detail and photorealistic facial representations.
Recently, Diffusion Models have revolutionized the capabilities of generative methods by surpassing the performance of GANs.
In this work, we present \ourname{}, a diffusion-based 3D facial avatar generative model.
Leveraging diffusion principles, our model accurately generates relightable facial avatars, utilizing an identity embedding extracted from an ``in-the-wild'' 2D facial image. 
The introduced multi-modal diffusion model is the first to concurrently output facial reflectance maps (diffuse and specular albedo and normals) and shapes, showcasing great generalization capabilities.
It is solely trained on an annotated subset of a public facial dataset, paired with 3D reconstructions.
We revisit the typical 3D facial fitting approach by guiding a reverse diffusion process using perceptual and face recognition losses. 
Being the first 3D LDM conditioned on face recognition embeddings,
\ourname{} reconstructs relightable human avatars,
that can be used as-is in common rendering engines,
starting only from an unconstrained facial image,
and achieving state-of-the-art performance. 
\end{abstract}

\section{Introduction}
\label{sec:intro}

\begin{figure*}
\begin{center}
\includegraphics[width=1\linewidth]
                  {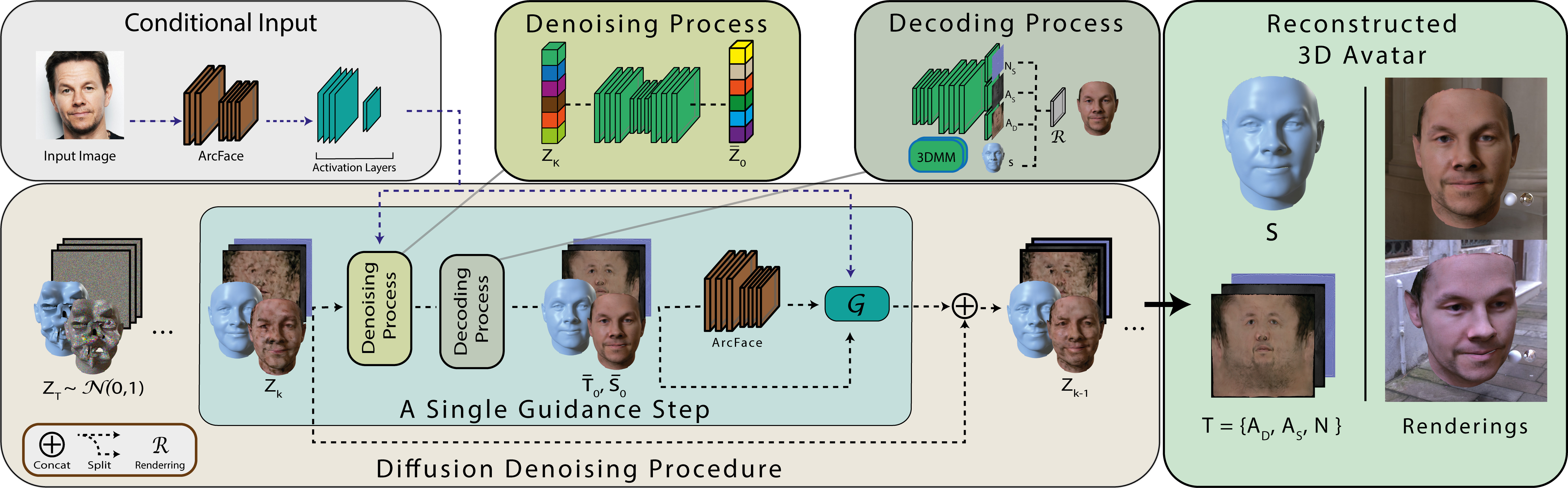}
\end{center}
  \caption{
  Overview of \ourname{}, a diffusion-based 3D facial generative network. Starting from Gaussian noise, our method generates facial avatars with relightable reflectance and shape, conditioned on an identity embedding. During sampling, a novel guidance algorithm ($\mathcal{G}$) is applied for further control of the resulting identity.
  $\mathbf{Z}_{T}, \mathbf{Z}_{k}$ and $ \mathbf{Z}_{k-1}$ are visualized in the actual picture space for illustration purposes.
  }
  \vspace{-0.25cm}
\label{fig:overview}
\end{figure*}

A fundamental objective of Computer Vision encompasses photorealistic 3D face reconstruction from a single image, which has gained significant attention from the research community over the past few decades. 
Its numerous applications in computer graphics, virtual reality, and entertainment, include avatar creation, face animation and manipulation~\cite{9010877, Taherkhani_2023_WACV, Gecer_2019_CVPR,athar2022rignerf,ma2023otavatar}. 
Despite the notable recent progress, accurate replication of personalized facial reconstructions continues to present a challenge. 
This is primarily due to the inherent ambiguity present in monocular images, associated with handling occlusions and capturing substantial variations in lighting conditions and facial expressions.
On top of that, captured 3D facial datasets are still relatively small~\cite{MICA:ECCV2022}, including biases and lacking generalization.

Since the introduction of the first 3D Morphable Model (3DMM)~\cite{Blanz1999AMM}, there has been tremendous progress in retrieving 3D facial shape information from a monocular image thanks to large-scale statistical models utilizing hundreds of subjects \cite{booth2018large,paysan20093d,FLAME:SiggraphAsia2017}.
More recently, Generative Adversarial Networks (GANs)~\cite{NIPS2014_5ca3e9b1}, and particularly the style-based generators~\cite{DBLP:journals/corr/abs-1812-04948,Karras2019stylegan2,Karras2020ada}, have demonstrated effectiveness in capturing intricate facial frequencies resulting into numerous subsequent studies~\cite{Gecer_2019_CVPR,gecer2021fast,Gecer_2021_CVPR}.
A fundamental challenge inherent in optimization fitting methods is their susceptibility to outliers, requiring to heuristically initialize the GAN’s $\mathbf{z}$ or $\mathbf{w}$ embedding for the back-propagation during inference to avoid instabilities~\cite{li2023robust}.
Additionally, they suffer from problems like unstable training and mode collapse~\cite{NIPS2016_8a3363ab,kodali2017convergence}.

A solution to the aforementioned problems is the integration of the recent emerging Diffusion Models (DMs)~\cite{ho2020denoising}. 
Taking inspiration from the thermodynamics~\cite{Sohl-Dickstein2015}, DMs define a $T$-length Markov Chain by gradually adding normally distributed noise to the data and learn to predict the input noise for each step $t \in \{1,2,\cdots,T\}$.
This methodology has already been applied to synthesize and manipulate facial images~\cite{zhang2023dreamface,wang2022rodin, Paraperas_2023_ICCV}.
Nevertheless, most of these methods employ a conditional mechanism reliant on textual descriptions or other auxiliary information~\cite{Radford2021LearningTV}, thereby directing their attention not exclusively towards the faithful reconstruction of the input facial identity and the exploration of the potential of robust identity embeddings. Moreover, Relightify~\cite{Paraperas_2023_ICCV} requires partially completed facial UV maps and third-party extracted facial shapes as input, thus being prone to these third-party failures.

This study presents \ourname{}, the first multi-modal diffusion model that synthesizes high-fidelity facial avatars given only an input facial image, starting from a randomly initialized Gaussian noise.
By harnessing the impressive generation capabilities of LDMs~\cite{Rombach_2022_CVPR}, we delve into their potential in 3D facial reconstruction.
Our approach enables the synthesis of facial avatars by combining facial UV reflectance maps and facial geometry while utilizing an identity embedding layer as a conditioning mechanism.
The diffusion process is applied to the concatenation of the latent vectors of the facial texture maps and shape, while a VQGAN AutoEncoder~\cite{Esser_2021_CVPR} and 
a 3DMM model (LSFM~\cite{booth2018large}) are used to decode them. 
Moreover, we present a novel facial guidance algorithm incorporated into the reverse diffusion process for accurately reconstructing a target facial identity.
\ourname{} generates high-fidelity facial avatars while achieving a state-of-the-art identity preservation score.
The training of our model involves fitting a facial reconstruction network~\cite{lattas2023fitme} to a manually selected and curated set of images acquired from the CelebA-HQ dataset~\cite{karras2018progressive}.
The fitting process results in acquiring facial shape and facial reflectance maps.
Overall, in this paper:
\begin{itemize}
\setlength\itemsep{0em}
    \item We introduce \ourname{},
    a multi-modal diffusion-based generative model that jointly produces facial geometry and appearance. The facial appearance consists of diffuse albedo, specular albedo, and normal maps, enabling photorealistic rendering.
    \item We show the first diffusion model conditioned on identity embeddings, acquired from an off-the-shelf face recognition network, whilst introducing a SPADE~\cite{park2019SPADE}-conditioned UNet architecture.
    \item We present unconditional samples of relightable avatars,
    but most importantly, we achieve facial reconstruction from a single ``in-the-wild'' image through identity embedding conditioning and guidance.
\end{itemize}

\section{Related Work}
\begin{figure*}
\begin{center}
\includegraphics[width=1\linewidth]{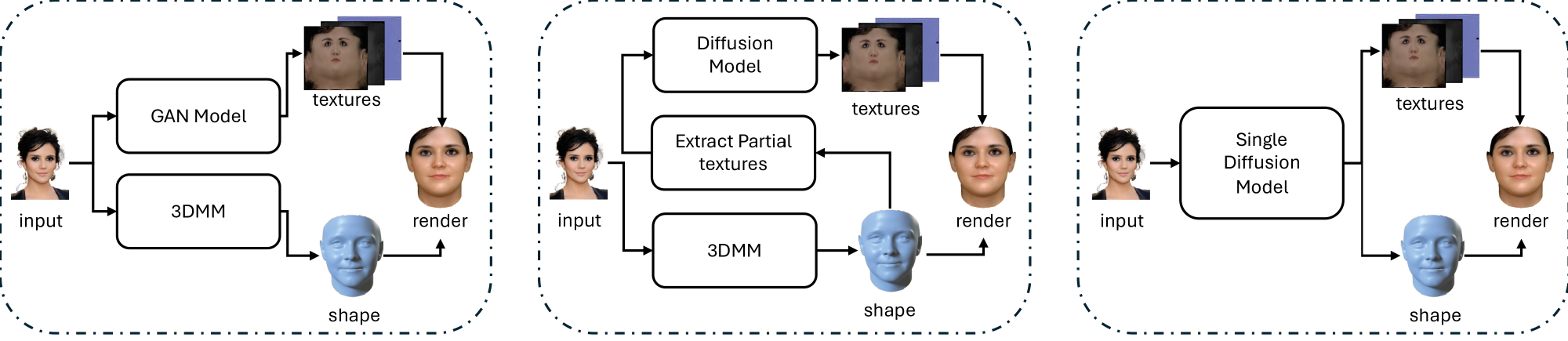}
\begin{scriptsize}
\makebox[0.332\linewidth][c]{\textbf{a)} FitMe \cite{lattas2023fitme}}\hfill
\makebox[0.332\linewidth][c]{\textbf{b)} Relightify \cite{Paraperas_2023_ICCV}}\hfill
\makebox[0.332\linewidth][c]{\textbf{c)} \ourname{} (Ours) }\hfill
\end{scriptsize}
\vspace{-0.5cm}
\end{center}
  \caption{Differences with existing state-of-the-art methods \cite{lattas2023fitme, Paraperas_2023_ICCV}:
  Prior works rely on multiple separate models, which can fail on challenging inputs (Fig.~\ref{fig:Relightify}). 
  In contrast, our method uses only a single Latent Diffusion Model for both shape and reflectance texture prediction, achieving simplified architecture, training, and robustness.
  }
  \vspace{-0.5cm}
\label{fig:method_comp}
\end{figure*}

\subsection{Face Modeling}
Various works followed the introduction of the emblematic first 3D Morphable Model (3DMM)~\cite{Blanz1999AMM} trained on 200 distinct subjects. 
Numerous works have since proposed improvements, especially on realistic expressions ~\cite{cao2013facewarehouse,yang2011expression,FLAME:SiggraphAsia2017,breidt2011robust,amberg2008expression,li2010example,thies2015real,bouaziz2013online,9156669}, large-scale dataset and model releases ~\cite{paysan20093d,booth20163d,smith2020morphable,dai20173d}.
However, those models could not capture high-frequency details due to their linear nature. 
To deal with this, many studies integrated 3DMMs with Deep Neural Networks~\cite{tewari2019fml,tran2019learning}, Mesh Convolutions~\cite{moschoglou20193dfacegan}, GANs~\cite{Gecer_2019_CVPR,gecer2021fast,gecer2020synthesizing} and VAEs~\cite{bagautdinov2018modeling,lombardi2018deep,wei2019vr,ranjan2018generating}.
Additionally, the photorealistic rendering of implicit representation-based methods~\cite{mildenhall2020nerf,Park_2019_CVPR} led to numerous approaches ~\cite{Gao-portraitnerf, hong2021headnerf,yenamandra2021i3dmm,Galanakis_2023_WACV,Or-El_2022_CVPR, ma2023otavatar,athar2022rignerf}.
However, their applications and editability remain challenging, hence this work focuses on explicit representations.
Emphasizing the acquisition of highly detailed facial texture and faithful reflectance maps, Avatarme~\cite{Lattas_2020_CVPR} and AvatarMe++~\cite{lattas2021avatarme++} treat the facial texture as a combination of diffuse albedo, specular albedo and normals UV maps whereas ReflectanceMM\cite{han2023ReflectanceMM} models spatially varying BRDF, while trained in low-cost acquired data.
In a different vein, the authors of MICA~\cite{MICA:ECCV2022} leverage a robust face recognition network~\cite{deng2019arcface} to generate facial shapes. In \cite{wood2022dense}, dense facial landmarks were introduced for better shape reconstruction, whilst a transformer-based facial reconstruction network was introduced in \cite{tokenface}.
The techniques above and more recent methodologies~\cite{10.1145/3450626.3459936, lei2023hierarchical, Chai_2023_ICCV, rai2023towards}, which utilize displacement maps for finer facial shape details, consistently achieve state-of-the-art performance in facial shape competitions such as NoW~\cite{RingNet:CVPR:2019} and REALY~\cite{REALY}.
However, many of these approaches produce facial texture with baked illumination or fail to generate it.
Recent works such as ~\cite{lattas2023fitme, dib2021towards, Luo_2021_CVPR} aim to concurrently reconstruct facial shape and texture but still rely on separate dedicated models for each component (Fig.~\ref{fig:method_comp}).
Closer to our work is AlbedoGAN~\cite{rai2023towards}, which introduces a single network for acquiring both facial shape and texture. 
However, contrary to our approach, the proposed methodology generates maps containing baked illumination and cannot handle wearables.

\subsection{Diffusion Models}
Inspired by~\cite{Sohl-Dickstein2015,song2019generative,song2021scorebased,song2020improved}, the authors of~\cite{NEURIPS2021_49ad23d1} showed that DMs can perform better than the widely used GAN-based methods in image synthesis tasks. 
The high-quality samples and the more stable training have attracted the attention of the research community leading to a great variety of applications in image generation~\cite{vahdat2021score, sinha2021d2c, Rombach_2022_CVPR, DBLP:journals/corr/abs-2102-12092,ho2022classifier,Rombach_2022_CVPR}, text-to-image~\cite{nichol2021glide, Kim_2022_CVPR}, 
text-to-3D~\cite{poole2022dreamfusion}, Pointcloud~\cite{Zhou_2021_ICCV,zeng2022lion,Lyu2021ACP} and Mesh~\cite{Liu2023MeshDiffusion} generation.
Closest to our work, other diffusion-based facial models~\cite{zhang2023dreamface,wang2022rodin, Paraperas_2023_ICCV} were presented.
Unlike our single-stage methodology, the former two employ a coarse-to-fine approach to attain identity information. 
Furthermore, Rodin~\cite{wang2022rodin} uses an implicit representation~\cite{Chan_2022_CVPR} for facial shape but doesn't prioritize faithful reconstruction of the input facial identity.
On the contrary, \ourname{} integrates a fully controllable 3DMM model, facilitating the generation of high-quality facial avatars that accurately represent the input identity.
On the other hand, Relightify~\cite{Paraperas_2023_ICCV} uses an unconditional denoising network for filling partially completed facial UV maps extracted from third-party models.
Although it may seem similar to our proposed methodology, \ourname{} has several advantages over Relightify, which we discuss in detail in Section~\ref{sec:discusssions}.

\section{Method}

In this work, we propose \ourname{}, a latent-diffusion-based approach to reconstruct facial avatars. We harness the power of diffusion models for the generative and fitting process as, by nature, they are very robust in both processes since they directly operate on the image space \cite{Rombach_2022_CVPR}.
On the other hand, GANs suffer from various issues such as mode collapse during training \cite{thanh2020catastrophic} or unrealistic outputs in fitting methods, resulting in unnecessary heuristics \cite{lattas2023fitme} to stabilize the outputs at the expense of fidelity.

A facial avatar can be formulated as a combination of a mesh $\mathbf{S}$ and texture $\mathbf{T}$ which is defined as a combination of facial reflectance UV maps, namely diffuse albedo ($\mathbf{A}_{D}$), specular albedo ($\mathbf{A}_{S}$) and normals ($\mathbf{N}$).
Also, let us denote an ``in-the-wild'' image containing a face as $\mathbf{I}$, and $\mathbf{V}_{trgt}$ as the corresponding target identity embedding of the appearing face.
Given $\mathbf{V}_{trgt}$ as input, \ourname{} generates a 3D facial avatar of the same identity as the one in $\mathbf{I}$, 
alongside with the current scene's illumination parameters (ambient, diffuse, and specular lighting and lighting direction). An overview of our method is illustrated in Fig.~\ref{fig:overview}, whereas the rest of the section includes a detailed presentation of our method's architecture (Sec.~\ref{sec:model_architecture}),
the proposed conditional input (Sec.~\ref{sec:condition_input}), the training scheme  (Sec.~\ref{sec:training}), and finally the identity-guidance sampling procedure 
 (Sec.~\ref{sec:sampling}).

\subsection{Model Architecture} \label{sec:model_architecture}

\ourname{} is a latent-diffusion based approach~\cite{Rombach_2022_CVPR}. 
This is motivated by the challenges posed by the large number of parameters and the computational expenses associated with simultaneously generating multiple meshes and texture images.
Thus, we represent facial avatars as a latent vector containing latent information about the shape, facial texture, and scene illumination:
$\mathbf{z} = \{\mathbf{z}_{tex} | \mathbf{z}_{shp} | \mathbf{z}_{ill} \} \in \mathbb{R}^{4288}$,
where $\mathbf{z}_{tex} \in \mathbb{R}^{4096}$ signifies the facial reflectance latent vector, $\mathbf{z}_{shp} \in \mathbb{R}^{183}$ the latent shape vector and $\mathbf{z}_{ill} \in \mathbb{R}^{9}$ the scene illumination parameters.
The latent shape vector $\mathbf{z}_{shp}$ can be further separated into the identity parameters $\mathbf{z}_{shp_{i}} \in \mathbb{R}^{158}$ and expression parameters $\mathbf{z}_{shp_{e}} \in \mathbb{R}^{25}$.

\ourname{} is composed of a 3D statistical model $\mathcal{F}_{shp}$~(LSFM~\cite{booth2018large}) that generates facial geometry,
a branched multi-modal AutoEncoder~\cite{Esser_2021_CVPR,Rombach_2022_CVPR} that generates facial reflectance maps and a denoising UNet AutoEncoder~\cite{DBLP:journals/corr/RonnebergerFB15}.
For a set of identity $\mathbf{z}_{shp_{i}}$ and expression $\mathbf{z}_{shp_{e}}$ parameters, the PCA face model $\mathcal{F}_{shp}$ generates a facial mesh $\mathbf{S}$ following the formula:
\begin{equation*}
    \mathbf{S} = \mathcal{F}_{shp} (\mathbf{z}_{shp}) = \textbf{U}_i \cdot \mathbf{z}_{shp_{i}} + \textbf{U}_e \cdot \mathbf{z}_{shp_{e}} + \textbf{m}_i
\end{equation*}
where $\mathbf{U}_i$ and $\mathbf{U}_e$ are the identity and expression bases, respectively, and $\mathbf{m}_s$ is the mean face.
Additionally, we incorporate a robust branched  VQGAN~\cite{Esser_2021_CVPR,Rombach_2022_CVPR}, to function as a multi-modal texture UV generator. 
More specifically, the VQGAN encoder $\mathcal{E}$ concurrently encodes facial diffuse albedo $\mathbf{A}_D$, specular albedo $\mathbf{A}_S$ and normals $\mathbf{N}$ into the same latent vector $\mathbf{z}_{tex}$, whereas the VQGAN decoder $\mathcal{D}$ reconstructs them given the input latent vector $\mathbf{z}_{tex}$.
Finally, following common diffusion-based methods~\cite{ho2020denoising, NEURIPS2021_49ad23d1, Rombach_2022_CVPR}, we utilize a UNet AutoEncoder~\cite{DBLP:journals/corr/RonnebergerFB15} with self-attention~\cite{NIPS2017_3f5ee243} layers $e_{\theta}(x_t, t)$ conditioned to the input time step $t \in \{1,\ldots,T\}$. 
We train the UNet to predict the injected noise $\epsilon$, sampled from a standard normal distribution, i.e., $\epsilon \sim \mathcal{N}(\mathbf{0,1})$. The training details of our method are presented in Sec.~\ref{sec:training}.

\subsection{Conditional input}\label{sec:condition_input}

In the underlying UNet model, we integrate a powerful conditioning mechanism to effectively learn all the necessary identity information. 
An ideal identity embedding must contain low and high-frequency information to accurately reconstruct the desired facial avatar.
To acquire such an identity embedding, we employ a powerful identity recognition network~\cite{deng2019arcface}, to which we feed the input facial image.
Then, the resulting conditioning vector combines the last feature vector with the intermediate activation layers of the identity recognition network.

Let $\mathcal{C}^n$ be the $n$-th intermediate layer of the identity recognition network.
We extract the intermediate activation maps $\mathcal{C}^{2} \in \mathbb{R}^{128\times 28\times 28},
\mathcal{C}^{3} \in \mathbb{R}^{256\times 14\times 14},
\mathcal{C}^{4} \in \mathbb{R}^{512\times7\times7}$ and concatenate them channel-wise with the identity embedding $\mathbf{V} \in \mathbb{R}^{512}$, which is expanded spatially.
Because of the 2D nature of our identity embedding and following  \cite{Galanakis_2023_WACV}, we use SPADE layers \cite{park2019SPADE} as a conditioning mechanism to inject the conditioning vector into the intermediate layers of the UNet.
\def \var {1}
\begin{figure*}[!t]
\begin{center}
 \begin{subfigure}[b]{\var\textwidth}
      \includegraphics[width=\textwidth]{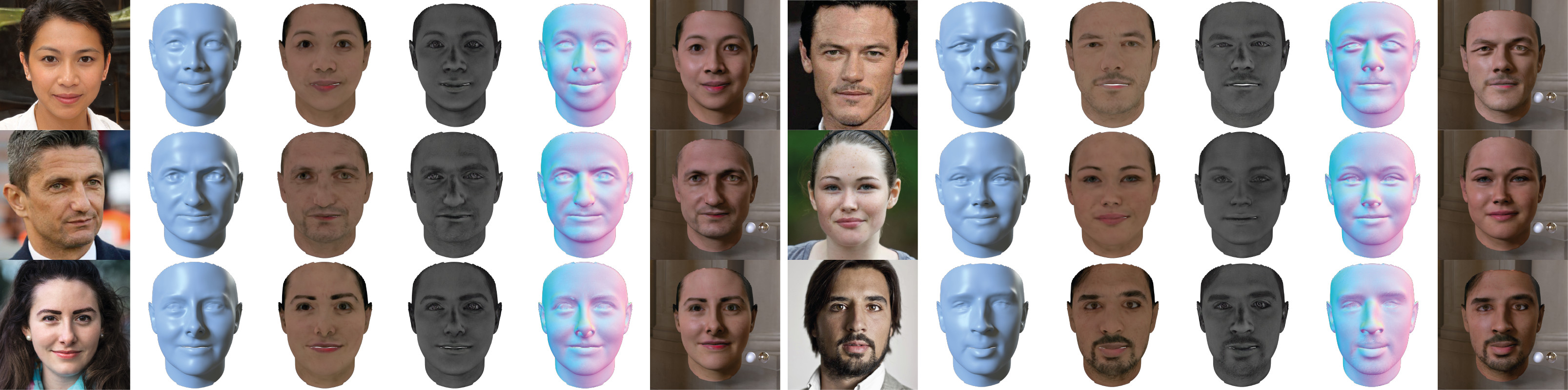}
         \begin{scriptsize}
             \makebox[0.088\linewidth][c]{Input}\hfill
             \makebox[0.081\linewidth][c]{Shape}\hfill
             \makebox[0.081\linewidth][c]{$\mathbf{A}_D$}\hfill
             \makebox[0.081\linewidth][c]{$\mathbf{A}_S$}\hfill
             \makebox[0.081\linewidth][c]{$\mathbf{N}$}\hfill
             \makebox[0.088\linewidth][c]{Rendering}\hfill
             \makebox[0.088\linewidth][c]{Input}\hfill
             \makebox[0.081\linewidth][c]{Shape}\hfill
             \makebox[0.081\linewidth][c]{$\mathbf{A}_D$}\hfill
             \makebox[0.081\linewidth][c]{$\mathbf{A}_S$}\hfill
             \makebox[0.081\linewidth][c]{$\mathbf{N}$}\hfill
             \makebox[0.088\linewidth][c]{Rendering}\hfill
         \end{scriptsize}
 \end{subfigure}
  \caption{
    Qualitative results of \ourname{}
    on ``in-the-wild'' facial images, 
    showing shape, reflectance, and environment map renderings.}
\label{fig:fittings}
\end{center}
\vspace{-0.7cm}
\end{figure*}

\subsection{Model Training}\label{sec:training}
Our training scheme consists of two phases: Initially, we conduct the training for the branched texture AE, followed by the subsequent training for the denoising UNet model.
The first part of our training protocol entails the training of the branched multi-modal AE~\cite{Esser_2021_CVPR}, whereby triplets are employed as input data:
\begin{equation*}
    x = \{ \mathbf{A}_D, \mathbf{A}_S, \mathbf{N} \}, ~~~
    \mathbf{A}_D, \mathbf{A}_S, \mathbf{N} \in  \mathbb{R}^{512\times512\times3}
\end{equation*}
where $\mathbf{A}_D$ represents the diffuse albedo, $\mathbf{A}_S$ the specular albedo and $\mathbf{N}$ the normals. 
Our approach includes a branched multi-modal  discriminator~\cite{lattas2023fitme} in combination with a perceptual loss~\cite{zhang2018perceptual} as the training loss.
The discriminator is a path-based discriminator~\cite{Esser_2021_CVPR} comprising two branches, accommodating their different statistics \cite{lattas2023fitme}. 
The first branch gets as input the concatenation of diffuse and specular albedos $\mathbf{A}_D \bigoplus \mathbf{A}_S$, whereas the second branch receives the normals $\mathbf{N}$.
Furthermore, we adhere to the default training parameters outlined in~\cite{Rombach_2022_CVPR}.
For a given triplet $x_k$, the encoder $\mathcal{E}$ projects $x_k$ into a latent representation $\mathbf{z}_{tex} = \mathcal{E}(x_k)$,
where $\mathbf{z}_{tex} \in \mathbb{R}^{h \times w \times c}$. Then, the latent vector $\mathbf{z}_{tex}$ is fed into the decoder $\mathcal{D}$, which produces a reconstructed output triplet $\bar{x}_k$.
We downsample the input texture UVs by a factor of $f = H/h = 512/64 = 8$, following the downsampling investigations in ~\cite{Rombach_2022_CVPR} and due to computational limitations. 
We use the LSFM~\cite{booth2018large} model for the shape decoder, pre-trained with $\sim$10k identities.

After training the texture AutoEncoder, we freeze its weight parameters and embark on training the identity-conditioned UNet.
During this phase, we employ multiple components including the identity embedding $\mathbf{V}_{tgt}$, the shape $\mathbf{z}_{shp}$, the distinct facial texture maps $\mathbf{A}_{D}, \mathbf{A}_{S},$ and $\mathbf{N}$, and the scene lighting $\mathbf{z}_{ill}$.
At each training step, the facial reflectance maps undergo an initial encoding by the pre-trained encoder $\mathcal{E}$, thereby yielding the latent texture vector $\mathbf{z}_{tex}$. 
Then, the input vectors are concatenated, i.e., $\mathbf{z}_0 = \{\mathbf{z}_{tex}|\mathbf{z}_{shp}|\mathbf{z}_{ill}\}$.
Let us denote $\mathbf{z}_t$ the noisy counterparts of $\mathbf{z}_0$, resulting from $t$ steps of noise injection.
The UNet network gets $\mathbf{z}_t$ as input and learns to predict the injected noise following:
\begin{equation}
\label{eq:noise}
    L_{noise} := \mathbb{E}_{\mathcal{E}(x), \epsilon~\sim ~\mathcal{N}(0,1),t} \left[ \parallel \epsilon - \epsilon_{\theta} \left( \mathbf{z}_t, t, \mathbf{V} \right)\parallel\right]
\end{equation}
where $\epsilon$ is the ground truth injected noise, $ \epsilon_{\theta}$ the 
predicted injected noise from $\mathbf{z}_t$,
$t$ the diffusion time step,
and $\mathbf{V}$ signifies the 2D identity embedding vector (Sec.~\ref{sec:condition_input}).

In addition to the primary loss function $\mathcal{L}_{noise}$, our training scheme integrates additional losses intended to enhance robustness, which are the identity losses~\cite{Gecer_2019_CVPR, lattas2023fitme} $\mathcal{L}_{id}, \mathcal{L}_{per}$ and the shape loss $\mathcal{L}_{verts}$. 
A detailed definition of those losses is provided in the supplemental material along with more training details.
It is important to note that these auxiliary losses do not apply to the latent variables.
Thus, in every training step, the initial latent vector $\bar{\mathbf{z}}_0$ is estimated by:
$\bar{\mathbf{z}}_0 = \frac{\mathbf{z}_t - \sqrt{1- \bar{\alpha}} \epsilon}{\sqrt{\bar{\alpha}_t}}$.
The vector $\bar{\mathbf{z}}_0$ is decoded into the estimated initial avatar. 
Firstly,  $\bar{\mathbf{z}}_0$ is split into the estimated initial latent texture vector $\bar{\mathbf{z}}_{0_{tex}}$,
latent shape vector $\bar{\mathbf{z}}_{0_{shp}}$ and scene parameters $\bar{\mathbf{z}}_{0_{ill}}$.
The first two vectors are fed into the decoder $\mathcal{D}$ and the PCA model $\mathcal{F}_{shp}$ respectively. 
In this way, the estimated initial facial reflectance maps $\bar{\mathcal{T}}_0$ and shape $\bar{\mathcal{S}}_0$ are retrieved.
Additionally, under the estimated initial scene illumination $\bar{\mathbf{z}}_{0_{ill}}$, we acquire the initial identity rendering $\bar{I}_0$ using a differentiable renderer~\cite{ravi2020pytorch3d} while using a differentiable multi-texture map shader as introduced in ~\cite{lattas2021avatarme++,lattas2023fitme} under a single directional light.
Overall, the conditional denoising model is trained using the following formula:
\begin{equation*}
\mathcal{L} = \mathcal{L}_{noise} + \mathcal{L}_{id} + \mathcal{L}_{per} + \mathcal{L}_{verts}
\end{equation*}
where $\mathcal{L}_{noise}$ is defined in Eq.~\ref{eq:noise}, $\mathcal{L}_{id}$ is the identity distance, $\mathcal{L}_{per}$ the identity perceptual loss and $\mathcal{L}_{verts}$ the shape loss.

Even though both the forward and the reverse diffusion processes can be described via stochastic differential equations in a continuous time~\cite{song2021scorebased},  they can also be applied in discrete time by choosing a very small step each time.
The selection of the appropriate number of diffusion steps is based on the premise that, in the final step, the input data should be completely converted into random noise.
We follow the training parameters proposed by the authors of \cite{Rombach_2022_CVPR}, and we choose $T = 1000$ while using a linear noise schedule. Furthermore, \ourname{} is trained following the Classifier-Free approach~\cite{ho2022classifier}, aiming to generate novel identities without any prior input.
During training and with a probability of $\mathcal{P}_{uncond}=0.1$, we randomly set the input identity embedding equal to zero by feeding the face recognition network~\cite{deng2019arcface} with an empty image. 


\subsection{Sampling Procedure}\label{sec:sampling}

DMs generate new samples by reversing the diffusion process, commencing from an initial random Gaussian noise.
In parallel, the authors of ~\cite{song2020denoising} introduced Denoising Diffusion Implicit Models (DDIMs), which are implicit probabilistic models~\cite{mohamed2016learning}.
They conduct a modified reverse diffusion process with fewer diffusion steps than those required during the vanilla DDPM sampling. 

In our method, we adopt the DDIM sampling technique and integrate it into our trained architecture to generate facial avatars.
We select the number of sampling steps to be $T=50$ for the DDIM sampling process.
Aiming to generate accurate photorealistic avatars, we employ a novel guidance algorithm alongside the conditional input.
Our approach is inspired by \cite{NEURIPS2021_49ad23d1}, in which the authors introduce a score corrector network conditioned on the diffusion step.
In our implementation, we incorporate a face recognition network  $\mathcal{C}$ ~\cite{deng2019arcface} as a score corrector alongside an off-the-shelf facial landmark detector $\mathcal{M}$~\cite{bulat2017far} and a perceptual loss~\cite{zhang2018perceptual}.
The guidance method can be implemented for the vanilla  DDPM~\cite{NEURIPS2021_49ad23d1} and the DDIM~\cite{song2020denoising} sampling techniques.
For the generation of intermediate images, we employ a differentiable renderer~\cite{ravi2020pytorch3d} with the modifications introduced in ~\cite{lattas2023fitme,lattas2021avatarme++} under a single directional light. 
The pseudo-code and a detailed presentation of the proposed guidance method are presented in the supplemental while the guidance formula is the following:
\begin{equation}
    \mathcal{G} = \mathcal{G}_{id}^{cos} +
    \lambda_1 \mathcal{G}_{id}^{per}  +
    \lambda_2 \mathcal{G}_{mse} +
    \lambda_3 \mathcal{G}_{lan} +
    \lambda_4 \mathcal{G}_{vgg}
\end{equation}
where $\mathcal{G}_{id}^{cos}$ denotes the cosine similarity between the identity vectors, $\mathcal{G}_{id}^{per}$ the identity perceptual similarity, $\mathcal{G}_{mse}$ the photometric loss, $\mathcal{G}_{lan}$ is the distance between the 3D facial landmarks extracted by using $\mathcal{M}$~\cite{bulat2017far} and $\mathcal{G}_{vgg}$ is the perceptual similarity using \cite{zhang2018perceptual}. 
Fig.~\ref{fig:fittings} showcases examples of our method being applied to ``in-the-wild'' images.

\def \var {1}
\begin{figure*}[t]
    \begin{subfigure}[b]{\var\textwidth}
         \centering
         \includegraphics[width=\textwidth]{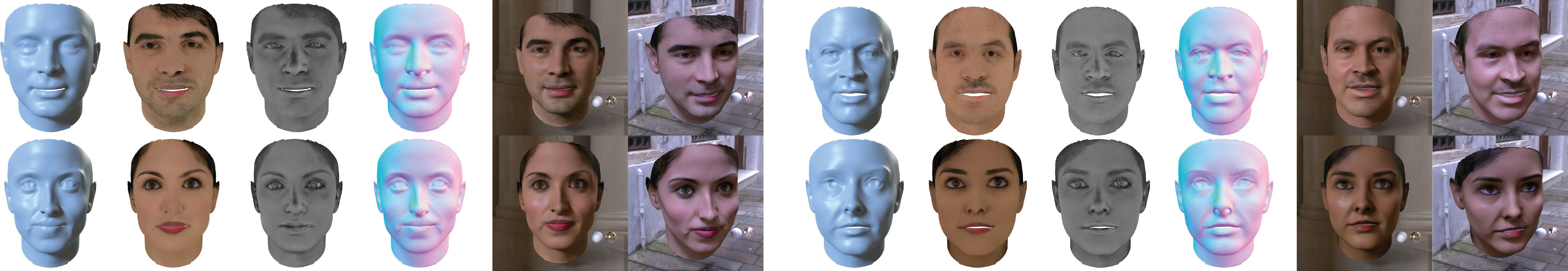}
         \begin{scriptsize}
             \makebox[0.075\linewidth][c]{Shape}\hfill
             \makebox[0.075\linewidth][c]{$\mathbf{A}_D$}\hfill
             \makebox[0.075\linewidth][c]{$\mathbf{A}_S$}\hfill
             \makebox[0.075\linewidth][c]{$\mathbf{N}$}\hfill
             \makebox[0.18\linewidth][c]{Rendering}\hfill
             \hspace{0.01\linewidth}
             \makebox[0.075\linewidth][c]{Shape}\hfill
             \makebox[0.075\linewidth][c]{$\mathbf{A}_D$}\hfill
             \makebox[0.075\linewidth][c]{$\mathbf{A}_S$}\hfill
             \makebox[0.075\linewidth][c]{$\mathbf{N}$}\hfill
             \makebox[0.18\linewidth][c]{Rendering}\hfill
         \end{scriptsize}
    \end{subfigure}
     \caption{ Samples generated by \ourname{} with unconditional sampling. Our method can generate diverse facial shapes and reflectance maps.}
     \vspace{-0.3cm}
     \label{fig:unconditional}
\end{figure*}

\section{Experiments}
\subsection{Dataset} \label{sec:sec_4_1}

Training such a diffusion model in a supervised manner
requires a large dataset of labeled sets of facial images $\mathbf{I}$,
facial textures $\mathbf{T}$, shape parameters $\mathbf{z}_{shp}$ and facial recognition embeddings $\mathbf{V}$.
Although a captured dataset could be used, there are no large enough public datasets~\cite{MICA:ECCV2022}.
As a workaround, we curate 9000 2D facial images from the CelebA-HQ Dataset \cite{karras2018progressive} $\mathbf{I}_i, i=0\dots{}9\cdot 10^{3}$,
on which we performed the following steps to acquire the labeled dataset:
A) We use a state-of-the-art face recognition model \cite{deng2019arcface},
to extract the identity latent embeddings $\mathbf{V}$,
which captures the facial structure with minimal interference from shading, age, and accessories.
B) We train a state-of-the-art StyleGAN-based \cite{Karras2019stylegan2} facial reconstruction network \cite{lattas2023fitme} $\phi$,
on public datasets of facial textures \cite{papaioannou2022mimicme, Lattas_2020_CVPR},
and use the LSFM 3DMM for the facial shape \cite{booth2018large}.
Following an iterative optimization \cite{lattas2023fitme}, 
we fit our model to the CelebA-HQ dataset~\cite{karras2018progressive} and acquire pseudo-ground truth facial textures, 3DMM shape weights, and scene illumination parameters 
$\phi(\mathbf{I}_i) \rightarrow \mathbf{A}_{D_i},\mathbf{A}_{S_i},\mathbf{N}_i, \mathbf{z}_{shp_i}, \mathbf{z}_{ill_i}$.
After manually filtering out all failed cases, we finalize a dataset consisting of paired images, facial reflectance textures, facial shape, scene illumination and latent vectors:
$\left\{ \mathbf{I}_i, \mathbf{A}_{D_i}, \mathbf{A}_{S_i}, \mathbf{N}_{i}, \mathbf{z}_{shp_i}, \mathbf{z}_{ill_i},\mathbf{V}_{i}\right\}$.

\subsection{Unconditional Sampling}

Following the classifier-free guidance (CFG)~\cite{ho2022classifier} training scheme, \ourname{}  can generate completely random facial identities without any prior input or supervision.
We present the unconditional generated diffuse albedos $\mathbf{A}_D$, specular albedos $\mathbf{A}_S$, normals $\mathbf{N}$, facial shapes $\mathbf{S}$ and renderings  in Fig.~\ref{fig:unconditional}.
This figure illustrates our method's ability to create distinct shapes and textures.

\subsection{Qualitative comparisons}

\def \var {0.5}
\begin{figure}[!t]
    \begin{subfigure}[b]{\var\textwidth}
         \centering
         \includegraphics[width=\textwidth]{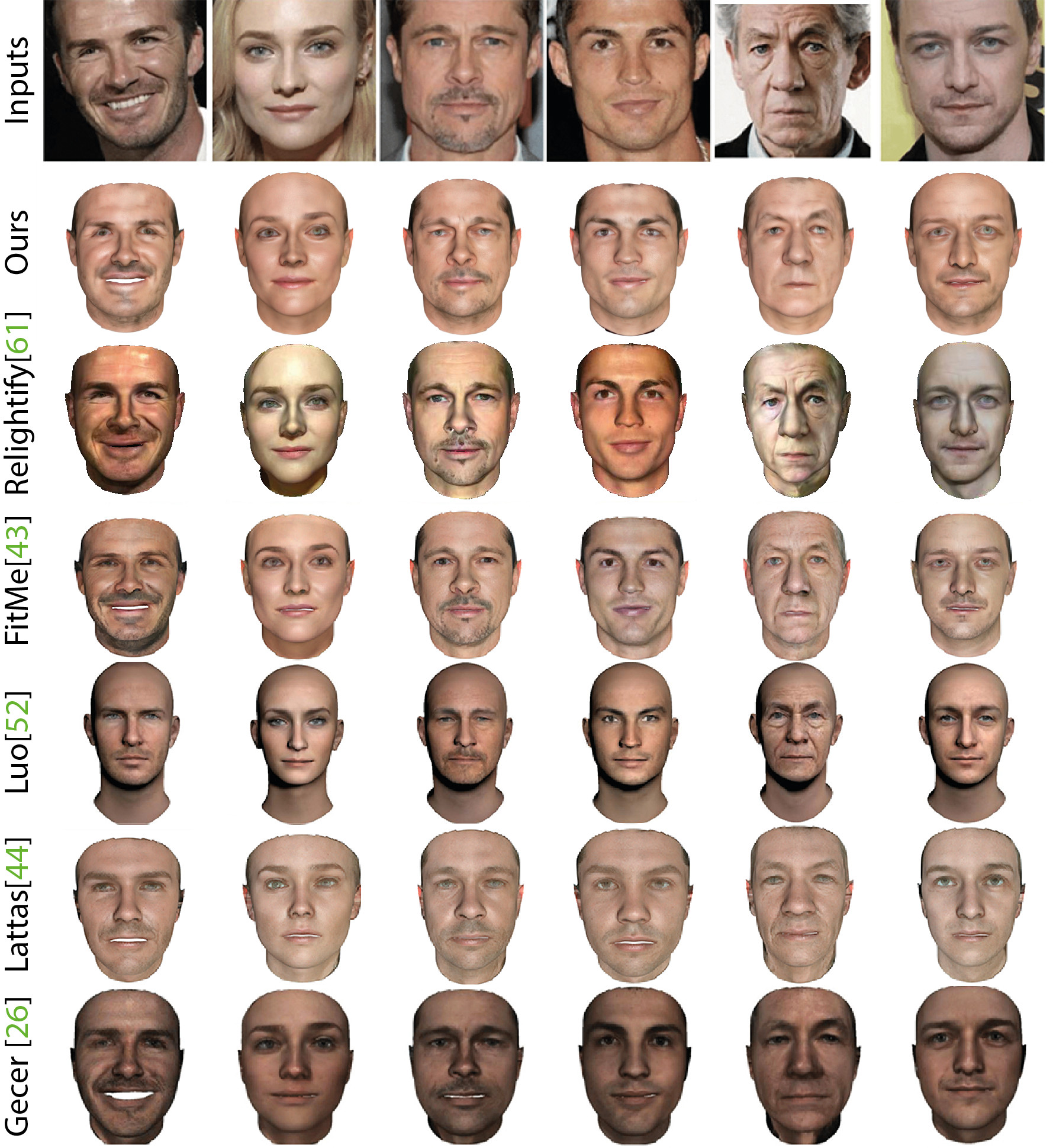}
         \label{}
    \end{subfigure}
    
     \caption{Qualitative comparison between \ourname{} and other monocular face reconstruction approaches~\cite{lattas2023fitme,Luo_2021_CVPR,lattas2021avatarme++,Gecer_2019_CVPR, Paraperas_2023_ICCV}. 
     }
    \label{fig:multi}
    \vspace{-0.3cm}
\end{figure}

We compare our method's generated samples with other monocular-image face reconstruction methods~\cite{lattas2023fitme,Luo_2021_CVPR,lattas2021avatarme++,Gecer_2019_CVPR, Paraperas_2023_ICCV} and present the generated samples in Fig.~\ref{fig:multi}.
Most of these techniques rely on GAN-based methods and employ fitting optimization procedures encompassing lighting, camera pose, and expression parameters, whereas Relightify~\cite{Paraperas_2023_ICCV} is the only diffusion-based approach.
Our method can capture finer details than most GAN-based methods(~\cite{lattas2023fitme, Luo_2021_CVPR,lattas2021avatarme++,Gecer_2019_CVPR}), whereas it generates equally detailed avatars like Relightify.

\begin{figure}[t]
    \centering
    \begin{subfigure}[t]{\linewidth}
        \centering
        \includegraphics[width=\linewidth]{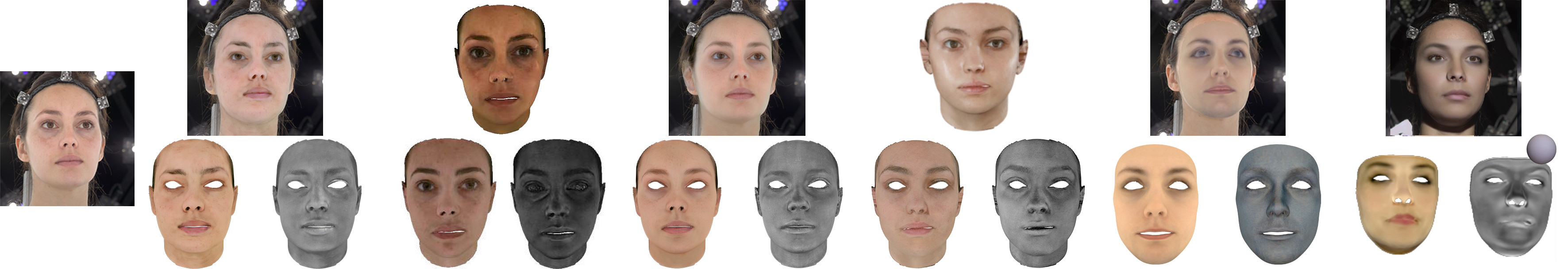}
        \begin{scriptsize}
        \makebox[0.07\linewidth][c]{Input}\hfill
        \makebox[0.155\linewidth][c]{Ours}\hfill
        \makebox[0.155\linewidth][c]{RF \cite{papaioannou2022mimicme}}\hfill
        \makebox[0.155\linewidth][c]{FM \cite{lattas2023fitme}}\hfill
        \makebox[0.155\linewidth][c]{AM \cite{lattas2021avatarme++}}\hfill
        \makebox[0.155\linewidth][c]{MM \cite{smith2020morphable}}\hfill
        \makebox[0.155\linewidth][c]{Dib \cite{dib2021towards}}\hfill
        \end{scriptsize}
        \caption{Comparison on Digital Emily \cite{alexander2010digital}.}
    \end{subfigure}%
    \\
    \vspace{0.1cm}
    \begin{subfigure}[t]{\linewidth}
        \centering
        \includegraphics[width=\linewidth]{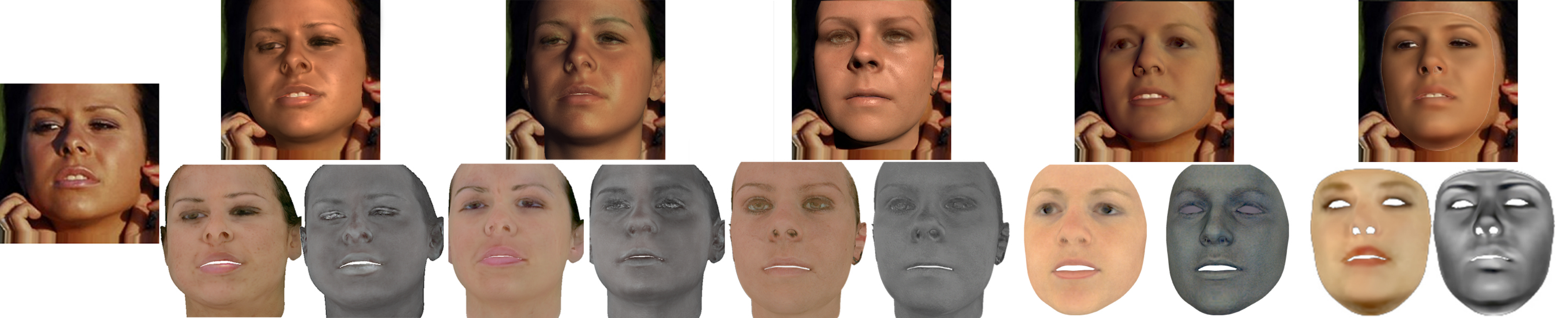}
        \begin{scriptsize}
        \makebox[0.10\linewidth][c]{Input}\hfill
        \makebox[0.18\linewidth][c]{Ours}\hfill
        \makebox[0.18\linewidth][c]{FM \cite{lattas2023fitme}}\hfill
        \makebox[0.18\linewidth][c]{AM \cite{lattas2021avatarme++}}\hfill
        \makebox[0.18\linewidth][c]{MM \cite{smith2020morphable}}\hfill
        \makebox[0.18\linewidth][c]{Dib \cite{dib2021towards}}\hfill
        \end{scriptsize}
        \caption{Comparison on a challenging case from Dib et al.~\cite{dib2021towards}.}
    \end{subfigure}
    \caption{
        Qualitative comparison on single-image reflectance acquisition
        against Relightify (RF) \cite{Paraperas_2023_ICCV}, FitMe (FM) \cite{lattas2023fitme}, AvatarMe++ (AM) \cite{lattas2021avatarme++},
        AlbedoMM (MM) \cite{smith2020morphable} and Dib et al.~\cite{dib2021towards}.
        Up: overlaid rendering, Left: diffuse, Right: specular.
    }
    \label{fig:emily_dib}
    \vspace{-0.4cm}
\end{figure}

\subsection{Quantitative comparisons}
\subsubsection{Facial Reflectance Acquisition Comparison}
We evaluate the quality of our method's generated facial reflectance maps by reconstructing 6 test subjects captured with a Light Stage~\cite{ghosh2011multiview}.
We compare the generated diffuse albedos $\mathbf{A}_D$, specular albedos $\mathbf{A}_S$ and normals $\mathbf{N}$ with the respective ground truth, and MSE, PSNR, and SSIM distances are measured. 
We compare our method's performance with AlbedoMM~\cite{smith2020morphable}, AvatarMe++~\cite{lattas2021avatarme++} FitMe~\cite{lattas2023fitme} and Relightify~\cite{Paraperas_2023_ICCV} and the results are presented in Tab.~\ref{tab:comp_ls} and Fig.~\ref{fig:emily_dib}.
\ourname{} generates state-of-the-art shape normals, whereas it performs on par with Relightify~\cite{Paraperas_2023_ICCV} in the diffuse and specular albedo scenarios.

\begin{table}[!t]
\centering
\setlength{\tabcolsep}{1.5pt}
\scriptsize
\begin{center}
\begin{tabular}{lccccccccc}
\toprule
        & \multicolumn{3}{c}{\textbf{Diffuse Albedo}}
        & \multicolumn{3}{c}{\textbf{Specular Albedo}}
        & \multicolumn{3}{c}{\textbf{Normals}} \\
        & $\downarrow$MSE       & $\uparrow$PSNR        & $\uparrow$SSIM 
        & $\downarrow$MSE       & $\uparrow$PSNR        & $\uparrow$SSIM  
        & $\downarrow$MSE       & $\uparrow$PSNR        & $\uparrow$SSIM\\
\midrule
\textbf{MM~\cite{smith2020morphable}}
        & 0.028                 & 15.82                 & 0.595
        & 0.007                 & 21.24                 & 0.608
        & -                 & -                 & -         \\
\textbf{AM~\cite{lattas2021avatarme++}}
        & 0.014                 & 18.30                 & 0.635
        & 0.005                 & 19.77                 & 0.640
        & 0.002                 & 27.26                 & 0.723         \\
\textbf{FM~\cite{lattas2023fitme}}       
        & 0.009                 & 21.12                 & 0.645 
        & 0.004                 & 23.95                 & 0.642 
        & 0.002                 & 26.77                 & 0.719         \\
\textbf{RF~\cite{Paraperas_2023_ICCV}} 
         & 0.009 & \textbf{22.47} & 0.647 &
         \textbf{0.003} & \textbf{27.17} & \textbf{0.710} & 
         0.002 & 26.69 & 0.719 \\ 
\textbf{Ours}                            
        & $\mathbf{0.009}$       & 21.16             & \textbf{0.647} 
        & 0.004        & 24.30             & 0.645
        & \textbf{0.001}        & \textbf{28.74}    & \textbf{0.734}\\
\bottomrule
\end{tabular}
\end{center}
\vspace{-0.3cm}
\caption{
    Quantitative comparison on 6 Light-Stage-captured data \cite{ghosh2011multiview},
    between our method, AlbedoMM \cite{smith2020morphable} (MM),
    AvatarMe++\cite{lattas2021avatarme++} (AM), FitMe\cite{lattas2023fitme} (FM) and Relightify\cite{Paraperas_2023_ICCV} (RF),
    measuring MSE, PSNR, and SSIM.
    Our method surpasses prior work in most cases or works on par with the current state-of-the-art in the rest.}
\label{tab:comp_ls}
\vspace{-0.5cm}
\end{table}

\subsubsection{Identity preservation experiment}\label{sec:identity_preservation}

\begin{figure}[!t]
 \centering
    \begin{subfigure}[t]{0.5\textwidth}
         \centering
         \includegraphics[width=\textwidth]{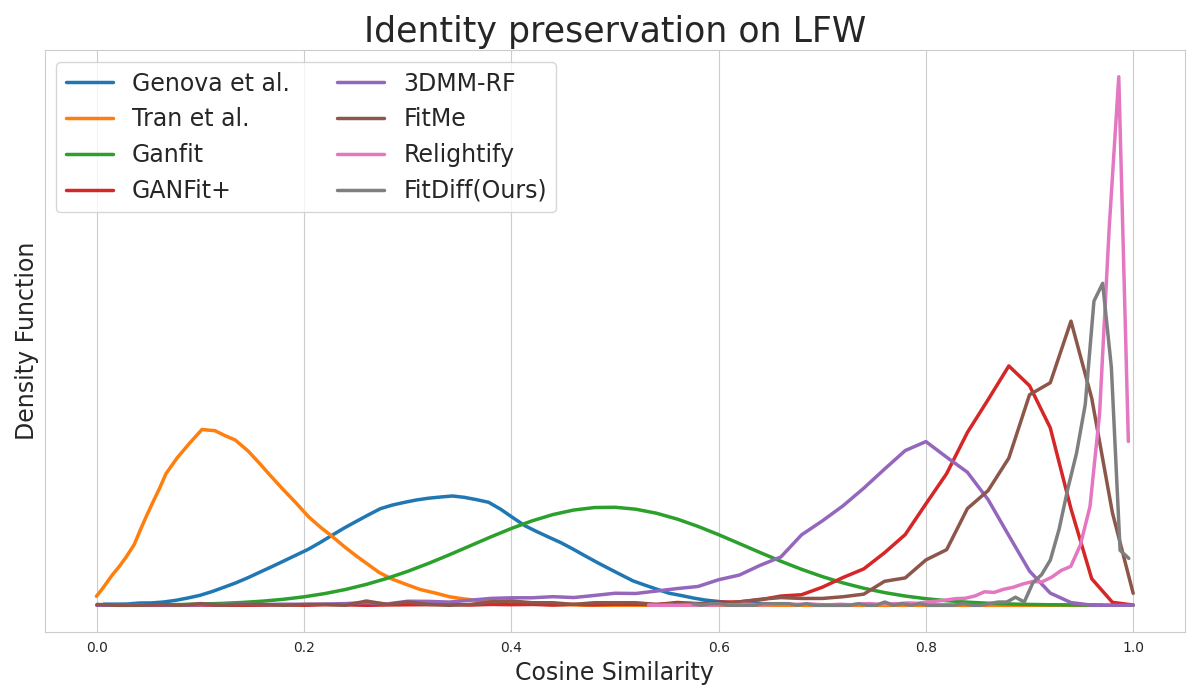}
    \end{subfigure}
     \caption{We compare our approach with ~\cite{Genova_2018_CVPR, tran2017regressing, Gecer_2019_CVPR, gecer2021fast, Galanakis_2023_WACV, lattas2023fitme, Paraperas_2023_ICCV}. \ourname{} performs on par with the current state-of-the-art method (Relightify~\cite{Paraperas_2023_ICCV}), while beating the rest.}
     \vspace{-0.6cm}
     \label{fig:exp_id_pers}
    
\end{figure}

One of the key elements of our model is the ability to generate the facial identity depicted in the provided ``in-the-wild'' image. 
We quantitatively measure this by conducting an identity preservation experiment~\cite{Genova_2018_CVPR,Gecer_2019_CVPR,gecer2021fast,Galanakis_2023_WACV}. 
We reconstruct the facial identities depicted in each image within the Labeled Faces in the Wild (LFW) dataset~\cite{LFWTech}. 
The reconstructed identities are fed into a face recognition network~\cite{BMVC2015_41} and the identity cosine distance is measured by comparing their activation layers.
As depicted in Fig.~\ref{fig:exp_id_pers},  \ourname{} outperforms the previous state-of-the-art face-reconstruction methods~\cite{lattas2023fitme,Gecer_2019_CVPR} and performs slightly less than current state-of-the-art method~\cite{Paraperas_2023_ICCV}.

\section{Ablation Study}

\subsection{Conditioning embedding}
The first ablation study focuses on the introduced conditioning embedding vector. Given the face recognition network~\cite{deng2019arcface}, We examine 3 different types of input identity embeddings a) using only the last layer b) using only the first 3 layers and c) using all four of them, as proposed in \ourname{}.  
We randomly pick about 100 ``in-the-wild" images found across the internet, and we fit them using \ourname{}.
During sampling, we don't use the guidance algorithm, as we want to let our network be entirely dependent on the identity embedding used. The first two rows of Table~\ref{tab:ablation} show that the proposed condition vector surpasses the other.

\subsection{Sampling guidance} 
Additionally, we examine the importance of the proposed facial guidance during the reverse diffusion process. 
Using the previously picked images, we consider three scenarios: a) sampling without any guidance, b) sampling using the classifier-free guidance~\cite{ho2022classifier} while using guidance scales $w=\{2, 9\}$, and c) our proposed method. 
The identity similarity scores are presented in Tab.~\ref{tab:ablation}, whereas visualizations of the generated examples are included in Fig.~\ref{fig:ablations}.
The guidance algorithm demonstrates superior reconstruction performance compared to alternative methodologies.

\subsection{Use of the conditioning mechanism}
\begin{figure}[!t]
    \centering
    \begin{subfigure}[b]{\linewidth}
         \centering
         \includegraphics[width=\textwidth]{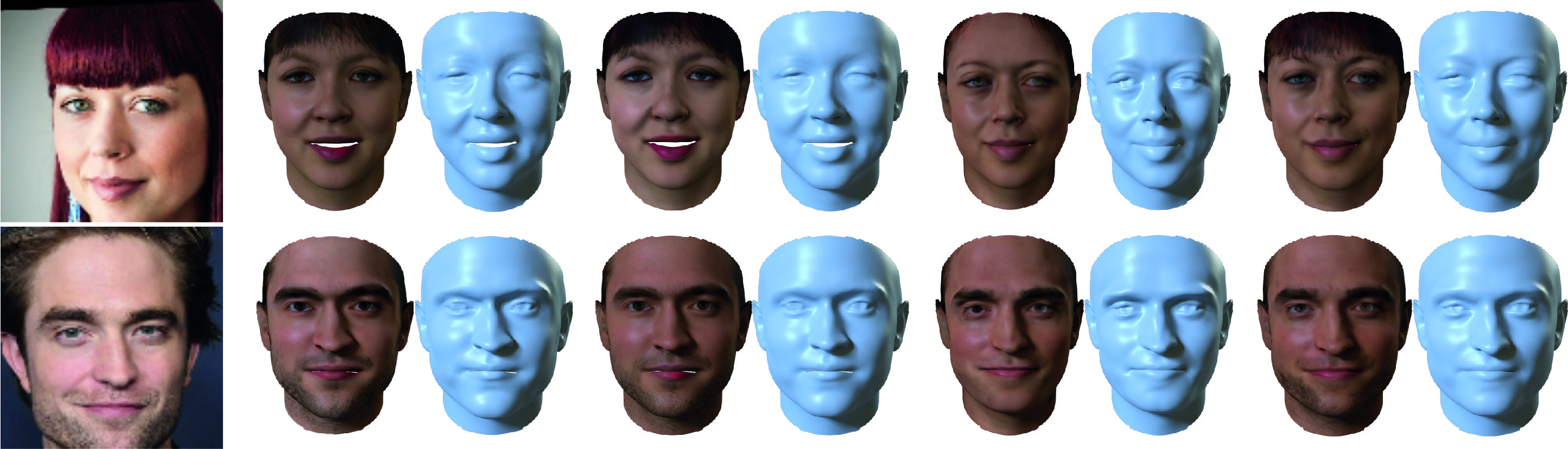}
        \begin{scriptsize}
        \makebox[0.135\linewidth][c]{\textbf{a)} Input}\hfill
        \makebox[0.25\linewidth][c]{\textbf{b)} w/o guid.}\hfill
        \makebox[0.2\linewidth][c]{\textbf{c)} CFG (w=2)}\hfill
        \makebox[0.2\linewidth][c]{\textbf{d)} w/o ident.}\hfill
        \makebox[0.2\linewidth][c]{\textbf{e)} Ours}\hfill 
        \end{scriptsize}
    \end{subfigure}
     \caption{Ablation study: From input images (a), we show results without the guidance algorithm (b), using CFG with w=2 (c), without identity embedding (d), and using our full method (e).}
     \label{fig:ablations}
     \vspace{-0.6cm}
\end{figure}

Another ablation study includes the necessity of our conditioning mechanism.
We compare the texture information between the generated samples with and without our conditioning mechanism, and examples of those are presented in Fig.~\ref{fig:ablations}d and~\ref{fig:ablations}e, respectively. 
These examples clearly show that finer details can be generated only when the corresponding identity embedding is used as input.

\begin{table}[!t]
\centering
\begin{scriptsize}
\setlength{\tabcolsep}{5pt}
\centering
\begin{tabular}{lccc}
\toprule
\textbf{Method}  & Using $n\in\{1,2,3\}$ layers & Using $n$ layer  & \ourname{} \\ 
\midrule
\textbf{ID Sim.} & 0.31 & 0.39 &   \textbf{0.45}  \\
\end{tabular}
\begin{tabular}{lccccc}
\toprule
\textbf{Method}  & Label Only & CFG (w=2)  & CFG (w=9)  & \textbf{Guidance}  \\ 
\midrule
\textbf{ID Sim.} &  0.43 & 0.49 &  0.45 &  \textbf{0.88}   \\
\bottomrule
\vspace{-0.3cm}
\end{tabular}
    \captionof{table}{Ablation study on the proposed conditioning input vector and the identity similarity performance of our method, using 
    with and without identity guidance.}
 \label{tab:ablation}
\end{scriptsize}
\vspace{-0.3cm}
\end{table}

\def \var {0.115}

\section{Discussion}\label{sec:discusssions}

\begin{figure*}[t]
    \begin{subfigure}[b]{0.5\linewidth}
    \includegraphics[width=\textwidth]{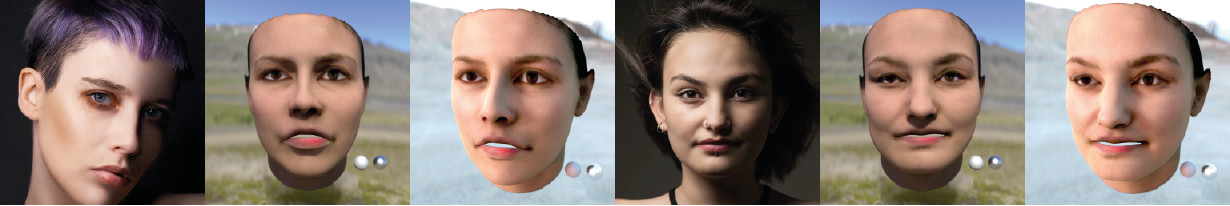}
    \begin{scriptsize}
             \makebox[0.15\linewidth][c]{\textbf{(a)} Input}\hfill
             \makebox[0.35\linewidth][c]{\textbf{(b)} Renderings}\hfill
             \makebox[0.15\linewidth][c]{\textbf{(c)} Input}\hfill
             \makebox[0.35\linewidth][c]{\textbf{(d)} Renderings}\hfill
         \end{scriptsize}
    \end{subfigure}
    \begin{subfigure}[b]{0.5\linewidth}
    \includegraphics[width=\textwidth]{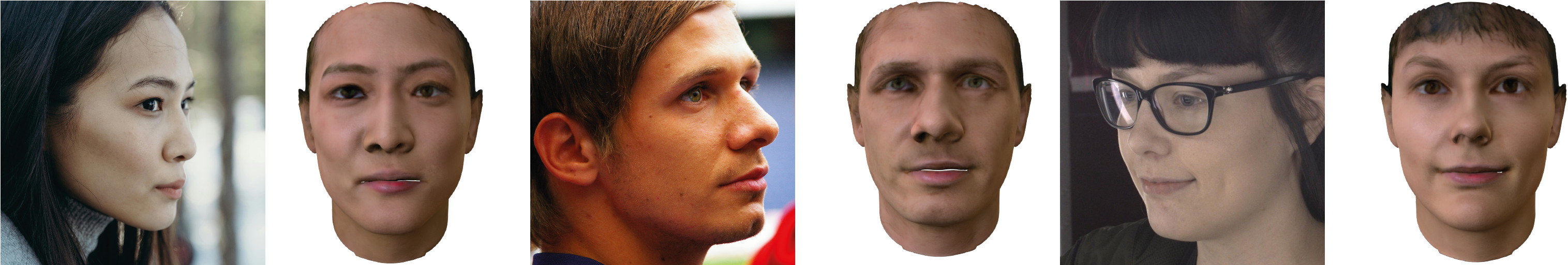}
    \begin{scriptsize}
             \makebox[0.18\linewidth][c]{\textbf{(a)} Input}\hfill
             \makebox[0.15\linewidth][c]{\textbf{(b)} Rendering}\hfill
             \makebox[0.20\linewidth][c]{\textbf{(c)} Input}\hfill
             \makebox[0.13\linewidth][c]{\textbf{(d)} Rendering}\hfill
             \makebox[0.20\linewidth][c]{\textbf{(c)} Input}\hfill
             \makebox[0.13\linewidth][c]{\textbf{(d)} Rendering}\hfill
         \end{scriptsize}
    \end{subfigure}
    \caption{Examples of \ourname{} under extreme illumination (left) and extreme angles (right)}
    \label{fig:chal-light}
    \vspace{-0.3cm}
\end{figure*}

\def \var {0.5}
\begin{figure}[!t]
    \begin{subfigure}[b]{\var\textwidth}
         \centering
         \includegraphics[width=\textwidth]{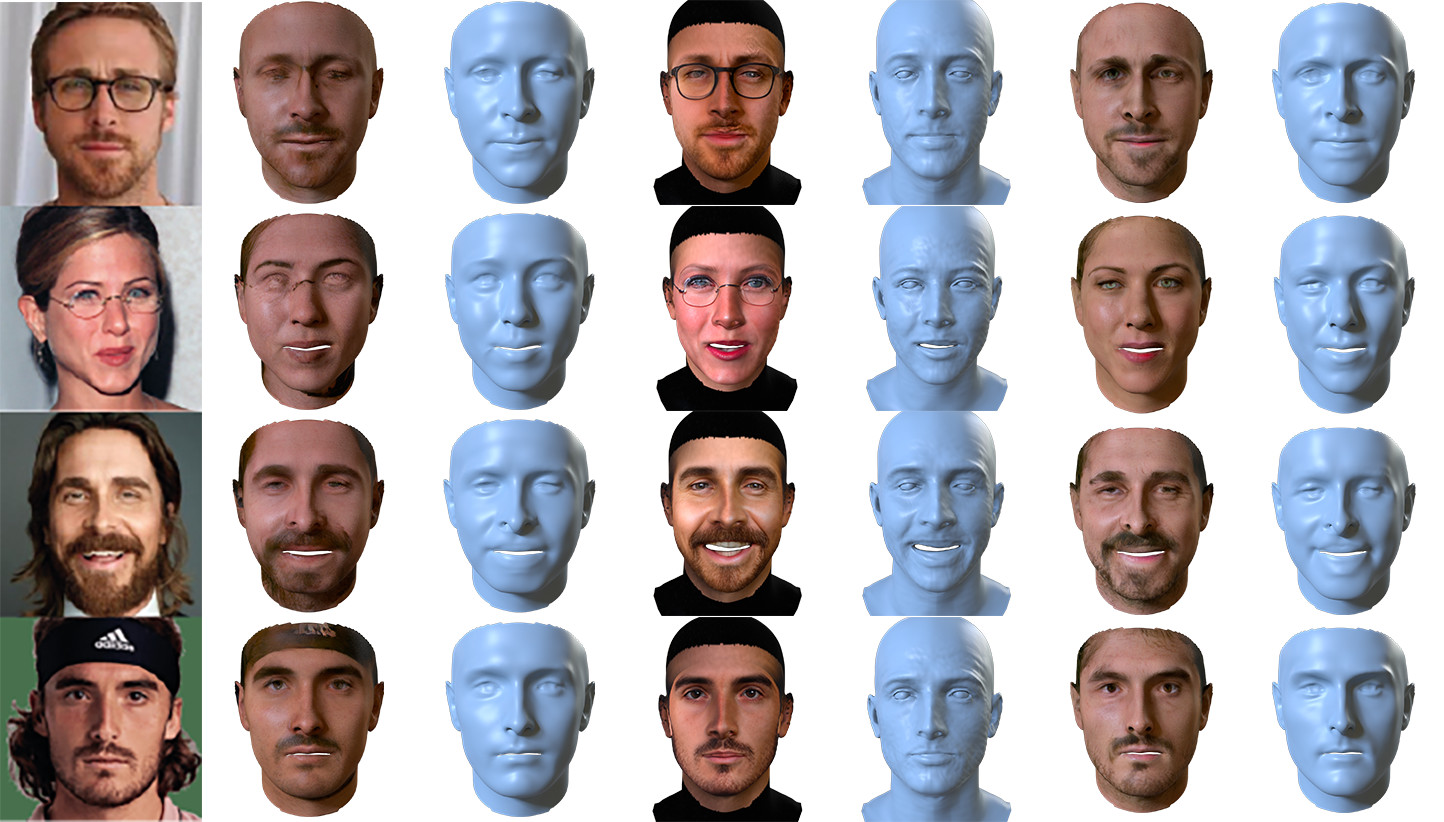}
         \begin{scriptsize}
             \makebox[0.142\linewidth][c]{\textbf{a)} Input}\hfill
             \makebox[0.285\linewidth][c]{\textbf{b)} Relightify~\cite{Paraperas_2023_ICCV}}\hfill
             \makebox[0.285\linewidth][c]{\textbf{c)} AlbedoGAN~\cite{rai2023towards}}\hfill
             \makebox[0.285\linewidth][c]{\textbf{d)} Ours}\hfill
         \end{scriptsize}
    \end{subfigure}
     \caption{Qualitative comparison between \ourname{}, Relightify~\cite{Paraperas_2023_ICCV}, and AlbedoGAN~\cite{rai2023towards} highlights the strengths of our approach. While Relightify is sensitive to occlusions and AlbedoGAN generates a single texture by replicating the input image, \ourname{} produces precise and high-quality facial reflectance maps.
     }
     \vspace{-0.25cm}
     \label{fig:Relightify}

     \def \var {1}
    \begin{center}
     \begin{subfigure}[b]{0.19\linewidth}
          \includegraphics[width=\textwidth]{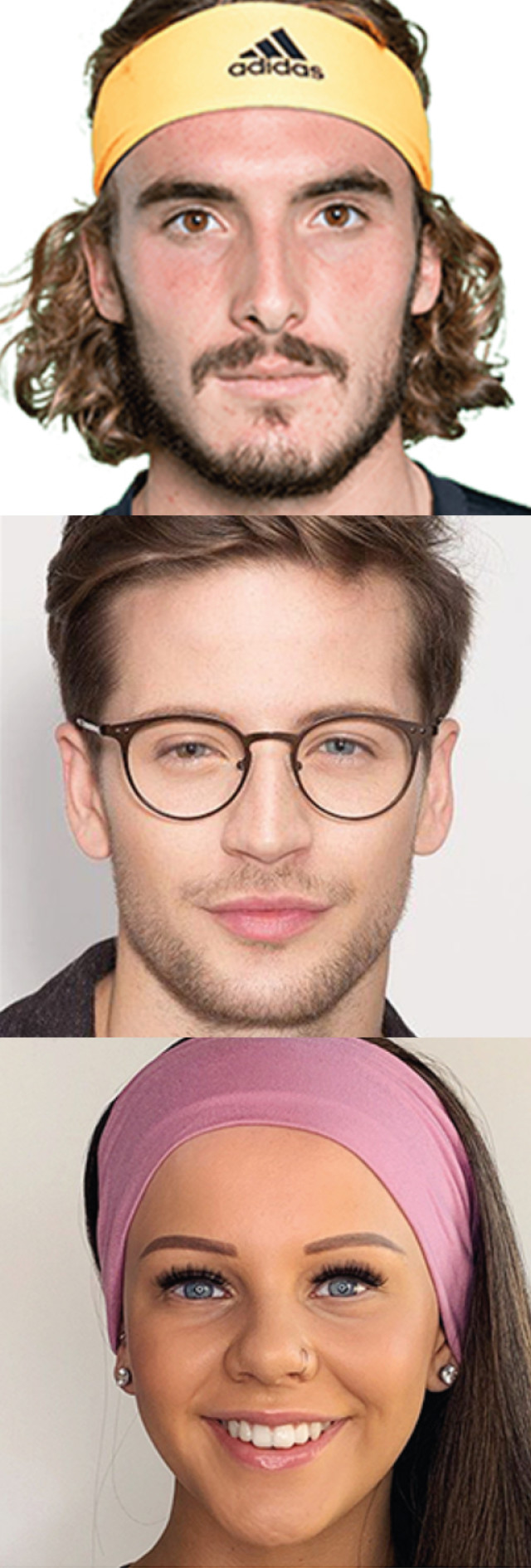}
          \caption{Input}
          \label{fig:fitme-compariso-a}
          \end{subfigure}
     \begin{subfigure}[b]{0.38\linewidth}
          \includegraphics[width=\textwidth]{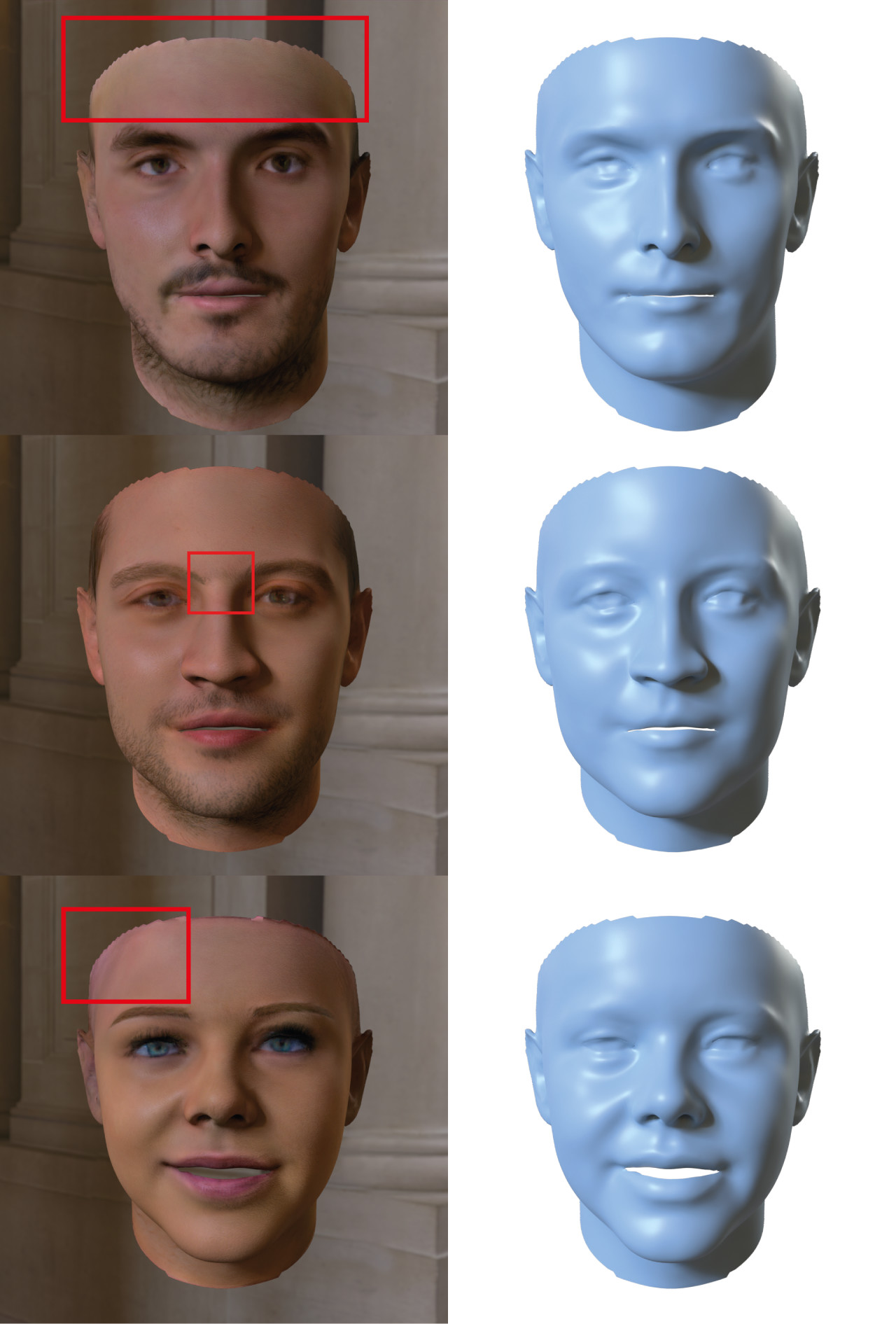}
          \caption{FitMe~\cite{lattas2023fitme}}
          \label{fig:fitme-compariso-b}    
          \end{subfigure}
     \begin{subfigure}[b]{0.38\linewidth}
          \includegraphics[width=\textwidth]{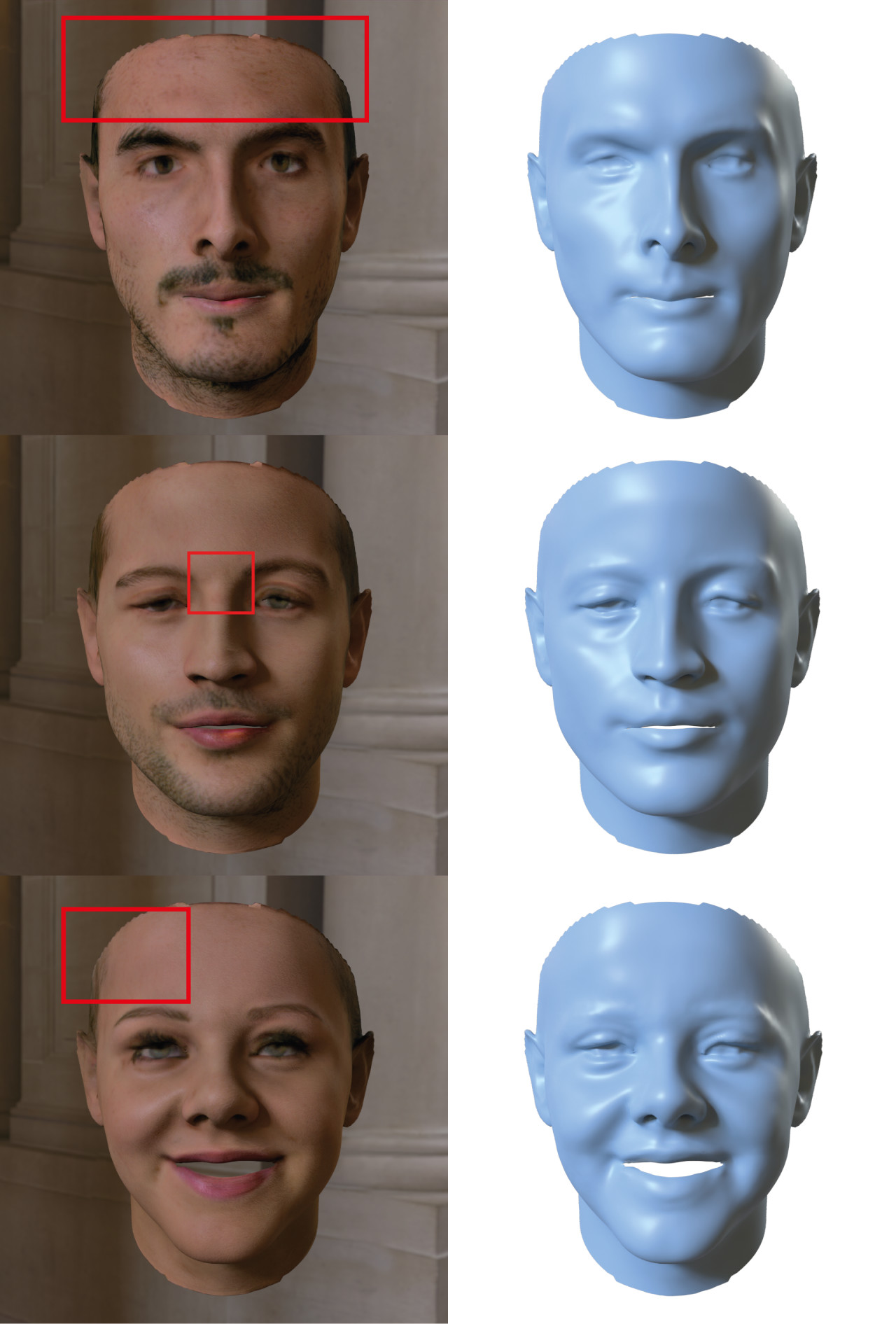}            \caption{\ourname{} (Ours) }
          \label{fig:fitme-compariso-c}
    
    \end{subfigure}
        \caption{
        We applied our method to challenging images from FitMe~\cite{lattas2023fitme}, provided by the authors. Despite occlusions in the input facial images, our reconstructions remain unaffected.}
        \label{fig:fitme-comparison}
        \end{center}
    \vspace{-0.7cm}
\end{figure}

\ourname{} is a latent diffusion model which concurrently carries out facial shape and texture generation as a combination of diffuse albedo, specular albedo, and normals.
A close work to ours is Relightify~\cite{Paraperas_2023_ICCV}, which only generates facial texture UV maps modeled like ours. 
It treats the texture reconstruction problem as an in-painting approach by copying the visible part of the facial textures acquired from third-party off-the-shelve approaches.
Then, a multi-modal diffusion model is used to complete the non-visible parts of the facial texture maps and reflectance. 
Relightify retains the visible input during inference, resulting in a great performance in identity preservation (Sec.~\ref{sec:identity_preservation}), which, however, introduces certain limitations:
a) low-resolution images result in low-resolution textures, and b) wearable items (glasses, headbands) partially leak into the reflectance textures, as shown in Fig.~\ref{fig:Relightify}.
On the contrary, \ourname{} concurrently generates facial texture maps and geometry from scratch and initializes the process from Gaussian noise.
This results in a) robust shape and texture reconstruction even under occlusions and accessories,
and b) always generating high-quality texture maps.
Despite these advantages, our approach performs slightly worse than Relightify in facial texture benchmarks.
This is due to Relightify's albedo-copying approach and the fact that \ourname{} was trained in sorely synthetic data, contrary to the light-stage training data used in Relightify. 
We sincerely believe that the robustness benefits above make \ourname{}
an important alternative approach with multiple important use cases.
In fact, our model can be used in tandem with Relightify to provide more accurate shape priors to further boost its performance.

On the other hand, the very recent AlbedoGAN~\cite{rai2023towards} is the first model that concurrently generates facial texture and shape. 
It provides FLAME~\cite{FLAME:SiggraphAsia2017} parameters alongside displacement maps for accurate facial shape reconstruction. 
While finetuning on the image input, it also produces a facial texture UV map, often incorporating baked illumination and accessories.
This consequently leads to reduced relightability in the generated avatars.
It also replicates the limitations observed in Relightify, where wearable items are incorporated into the final facial texture(Fig.~\ref{fig:Relightify}c).

Compared with FitMe~\cite{lattas2023fitme}, our experimentation results and qualitative assessments show that our generated texture maps achieve superior performance.  
This is due to the necessity in GAN-based fitting methods to meticulously initialize the GAN’s \textbf{z} or \textbf{w} embedding for back-propagation during inference to mitigate optimization instabilities.
The proposed method excels without the need for heuristic priors or regularization terms.


\vspace{-0.05cm}
\section{Conclusion}

In this paper, we introduced \ourname{}, a diffusion-based 3D facial generative model conditioned on identity embeddings from a pre-trained facial recognition system.
This approach captures diverse attributes like ethnicity, age, and gender from a single 2D image, with no restriction on quality, pose, or illumination. 
Our method jointly generates facial shapes, facial reflectance maps, and scene illumination parameters.
Through a series of experiments, \ourname{} showcases state-of-the-art performance in preserving identity and reconstructing facial reflectance, matching or surpassing established methods.
Finally, it can generate unconditional samples and handles occlussions, further highlighting its versatility and effectiveness. 
\paragraph{Acknowledgments:} S. Zafeiriou and part of the research was funded by the EPSRC Fellowship DEFORM (EP/S010203/1) and EPSRC Project GNOMON (EP/X011364/1).
{\small
\bibliographystyle{ieee_fullname}
\bibliography{main}
}

\end{document}


\title{Supplementary material: \ourname{}: Robust monocular 3D facial shape \\
 and reflectance estimation using Diffusion Models}


\author{Stathis Galanakis^{1,2}
\and
Alexandros Lattas^{1}
\and
Stylianos Moschoglou^{1}
\and
Stefanos Zafeiriou^{1}
\and
\\
$^1$Imperial College London \\
$^2$HUAWEI Noah's Ark Lab 
}

\maketitle

\section{Future Work}

In this work, we present the first-of-its-kind diffusion model conditioned on expressive facial embeddings, which is essentially a step towards a facial foundation model.
Such models require vast datasets of labeled data \cite{Rombach_2022_CVPR},
in our case paired facial images with geometry, reflectance
and identity embeddings.
These are immensely challenging to acquire in numbers,
and hence we have to rely on a synthetic dataset
and inherit its method's limitations \cite{lattas2023fitme}. 
Nevertheless, our method could be trivially extended to larger datasets of even scanned datasets (e.g.~\cite{yang2020facescape}),
given their availability.
Moreover, the ``fitting'' nature of our method,
is limited by the ambiguity between scene illumination and skin tone,
especially in single-image inference.
To that end, the recent method of TRUST~\cite{Feng:TRUST:ECCV2022},
could be incorporated into our diffusion model, as an additional conditioning mechanism, given however the availability of training data.

\begin{figure*}
\begin{center}
\includegraphics[width=1\linewidth]
                  {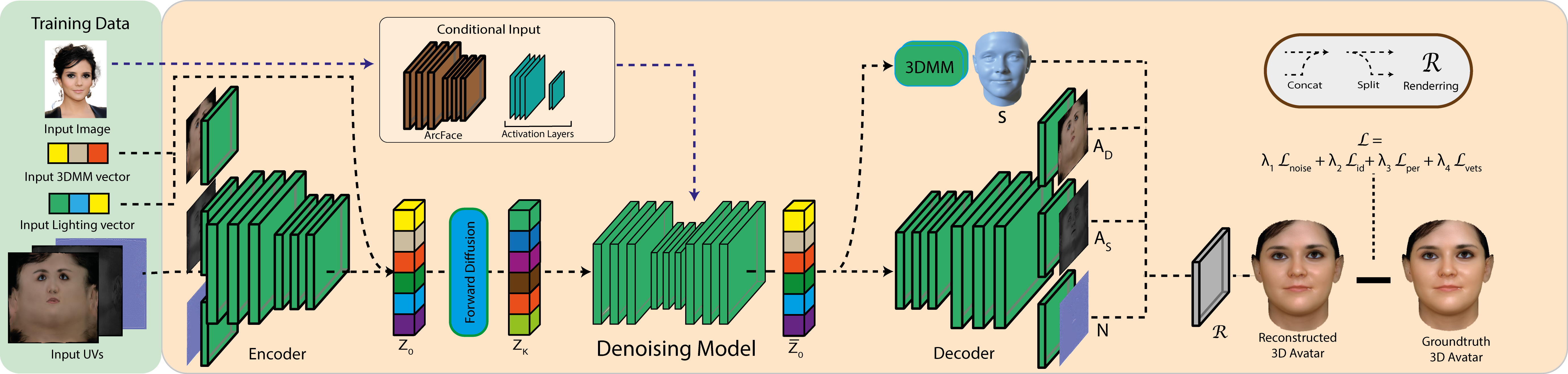}
\end{center}
  \caption{ Overview of the main phase of our training scheme: At each training iteration, the facial reflectance maps are first projected into the latent space and subsequently concatenated to the latent vector $\mathbf{z}_0$, to which noise is introduced. 
  After estimating the initial latent vector $\bar{\mathbf{z}}_0$ and rendering $(\mathcal{R})$ the estimated initial avatar,  perceptual and face recognition losses are applied.
  }
\label{fig:training}
\end{figure*}

\section{Implementation Details}
A comprehensive overview of this training approach is presented in  Fig.~\ref{fig:training}.
We provide the essential information required to reproduce our method.
The code-base for the brached multi-modal AutoEncoder is built on the public repository of the VQGAN AutoEncoder~\cite{Esser_2021_CVPR}.
We made the following changes:
A) The first downsampling layer of the encoder $\mathcal{E}$ and the last upsampling layer of the decoder $\mathcal{D}$ are branched, by making 3 copies of the respective layers. 
B) As proposed in FitMe~\cite{lattas2023fitme}, we use a branched discriminator, in the essence of having 2 copies of the main discriminator, except the last convolutional layer. The branch, dedicated for diffuse ($\mathbf{A}_D$) and specular ($\mathbf{A}_S$) albedos, gets a 6-channel input whereas the normals ($\mathbf{N}$) branch gets a 3-channel input.
\begin{table}[t]
    \centering
    \begin{tabular}{|c|c|c|c|}
        \hline
        \textbf{f} & $ |\mathcal{Z}|$  & \textbf{Embed.~dim~} \\
         8 & 16384 & 1 \\
         \hline
        \textbf{z channels} & \textbf{Channels} & \textbf{Channels mult.}\\
        4 &  128 & 1,2,2,4 \\
        \hline
        \textbf{Res.~Blocks} & \textbf{Attention Res.} & \textbf{Batch Size} \\
        2 & 32 & 16  \\
        \hline
    \end{tabular}
    \caption{Hyper-parameters used during training the branched multi-modal AutoEncoder.}
    \label{tab:vqgan}
\end{table}

On the other hand, the main training phase is built on the public repository of Latent Diffusion Models~\cite{Rombach_2022_CVPR}.
We modified the provided UNet code by turning it into a 1-D UNet network and replaced the attention-based conditional mechanism with SPADE layers~\cite{park2019SPADE}.
The hyper-parameters for the Conditional UNet using  SPADE layers are presented in Tab.~\ref{tab:hyper} while it gets trained for 800 epochs.

As mentioned in the main paper, the overall training loss under which \ourname{} is trained, is the following:
\begin{equation*}
\mathcal{L} = \mathcal{L}_{noise} + \mathcal{L}_{id} + \mathcal{L}_{per} + \mathcal{L}_{verts}
\end{equation*}
where $\mathcal{L}_{noise}$ is the noise prediction loss as defined in Section 3.3, $\mathcal{L}_{id}$ is the identity distance, $\mathcal{L}_{per}$ the identity perceptual loss and $\mathcal{L}_{verts}$ the shape loss.
\paragraph{Identity distance}
To supervise the identity similarity between the ground truth and the predicted facial avatars, we follow the methodology presented in \cite{Gecer_2019_CVPR, lattas2023fitme}. We employ a face recognition network~\cite{deng2019arcface} with $n$ layers : $\mathcal{C}^{n} (\mathbf{I}) : \mathbb{R}^{H\times W \times C} \rightarrow \mathbb{R}^{512} $. The identity distance is estimated by computing the identity similarity between the  feature emebeddings of the input image $\mathbf{I}$, and the estimated initial image $\bar{\mathbf{I}}_{0}$:
\begin{equation}
    \mathcal{L}_{id} = 1 - \frac{\mathcal{C}^{n} ( \bar{\mathbf{I}}_{0} ) \cdot \mathcal{C}^{n} ( \mathbf{I} ) }{\parallel \mathcal{C}^{n} ( \bar{\mathbf{I}}_{0} ) \parallel_2 \cdot \parallel \mathcal{C}^{n} ( \mathbf{I} ) \parallel_2}
\end{equation}
\paragraph{Identity perceptual loss}
To enforce perceptual consistency between the generated and ground truth avatars, we also penalize the discrepancy between the intermediate activation layers of the face recognition network $\mathcal{C}$. 
The identity perceptual loss is computed as:
\begin{equation}
    \mathcal{L}_{per} = \sum_{j}^{n} \frac{\parallel \mathcal{C}^{j} ( \bar{\mathbf{I}}_{0} ) - \mathcal{C}^{j} ( {\mathbf{I}} ) \parallel_2 } { H_{\mathcal{C}_{j}} \cdot W_{\mathcal{C}_j} \cdot C_{\mathcal{C}_j} }
\end{equation}

where $H_{\mathcal{C}_{j}}$, $W_{\mathcal{C}_{j}}$, and $C_{\mathcal{C}_{j}}$ denote the height, width, and number of channels of the $j$-th activation map, respectively.

\paragraph{Shape loss}
The difference between the estimated facial shape $\bar{\mathbf{v}}_{0}$ and the ground truth facial shape $\mathbf{v}$  is calculated using the L1-norm:
\begin{equation}
    \mathcal{L}_{verts} = \parallel  \bar{\mathbf{v}}_{0} - \mathbf{v} \parallel_1
\end{equation}

\begin{table}[t]
    \centering
    \begin{tabular}{|c|c|c|}
         \hline
         \textbf{Diffusion steps} & \textbf{Noise Schedule} & \textbf{Input Channels} \\
         1000 & linear & 1  \\ 
         \hline
          \textbf{Channels} & \textbf{Cond.~Dim} & \textbf{SPADE dim.} \\
          192 & 1048 & 128  \\
         \hline
          \textbf{Channels mult} & \textbf{Depth} &  \textbf{Heads}\\
         1,2,4,8 & 2  & 4 \\
         \hline
         \textbf{Heads Channels} & \textbf{Batch size} & \textbf{LR} \\
         32 & 16 & 3.2e-05 \\
         \hline
          
    \end{tabular}
    \caption{Hyper-parameters of the main training phase.}
    \label{tab:hyper}
\end{table}

\section{Guidance Algorithm}
\begin{figure}[!h]
    \centering
    \includegraphics[width=0.9\linewidth]{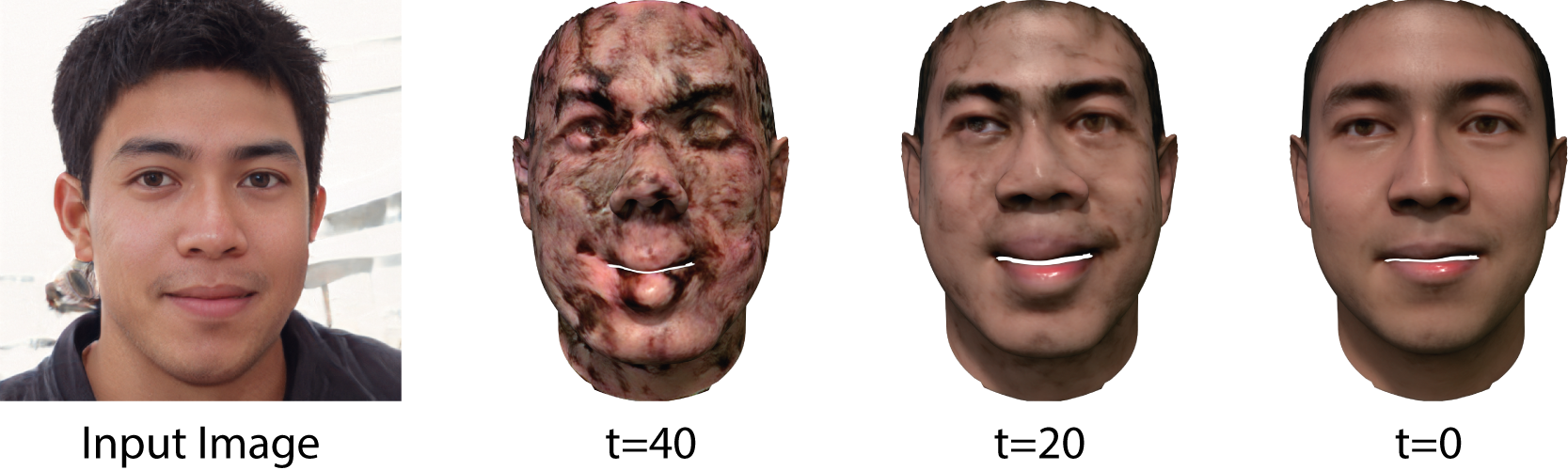}
    \caption{An example of the sampling process for $t=\{40, 20, 0\}$,}
    \label{fig:diffusion_process}
\end{figure}

\ourname{} is a diffusion-based architecture conditioned on an identity embedding vector.
It accurately generates facial identities by incorporating an effective identity guidance method during the sampling phase.
An example of this process is illustrated in Fig.~\ref{fig:diffusion_process}.
The proposed guidance method uses the guidance loss which is formulated as:
\begin{equation}
    \mathcal{G} = \mathcal{G}_{id}^{cos} +
    \lambda_1 \mathcal{G}_{id}^{per}  +
    \lambda_2 \mathcal{G}_{mse} +
    \lambda_3 \mathcal{G}_{lan} +
    \lambda_4 \mathcal{G}_{vgg}
\end{equation}
The values of the used lambdas are $\lambda_1=50, \lambda_2=10, \lambda_3=200$, $\lambda_4=1$ where we use a gradient scale $s=75$.
We run our sampling method for $T=50$ steps. 
When run on an NVIDIA Tesla V100-PCIE-32GB GPU, the diffusion sampling process takes about 54 seconds, timed comparable with other fitting methods like FitMe~\cite{lattas2023fitme} and Relightify~\cite{papantoniou2023relightify} which take about 50 seconds and about 1min respectively.


\subsection{Sampling Guidance pseudo-code}
\begin{algorithm}[!h]
\caption{Diffusion sampling using Identity Guidance}\label{alg:ddpm}
\begin{algorithmic}[1]
\Require A facial ``in-the-wild'' image $\mathbf{I}$, a gradient scale $s$, and networks $\mathcal{C}$\cite{deng2019arcface}, 
$\mathcal{M}$~\cite{bulat2017far},$\mathcal{F}_{shp}$\cite{booth2018large}, $\mathcal{V}$\cite{zhang2018perceptual}, and the multi-modal decoder $\mathcal{D}$ .
\Ensure $\mathbf{z}_0 = \{ \mathbf{z}_{tex}|\mathbf{z}_{shp}| \mathbf{z}_{ill} \}$. 
\State $\mathbf{z}_T = \{ \mathbf{z}_{{tex}_{T}}| \mathbf{z}_{{shp}_T}| \mathbf{z}_{{ill}_T}\} ~ \backsim ~ \mathcal{N}(\textbf{0},\textbf{1})$   
\State $\textbf{V}_{trgt} = \mathcal{C}(\mathbf{I})$
\ForEach{ t \textbf{from} T to 1} 
\State $\mathbf{\mu}, \mathbf{\Sigma} \leftarrow \epsilon_{\theta} (\mathbf{z}_t, t,\textbf{V}_{trgt})$
\State $\bar{\mathbf{z}}_0 = \frac{\mathbf{z}_t - \sqrt{1-\bar{\alpha_t}} \epsilon_{\theta} (\mathbf{z}_t, t,\textbf{V}_{trgt})}{\sqrt{\bar{\alpha_t}}}$
\State $\bar{\mathbf{I}}_0 \xleftarrow{\text{render}} \bar{\mathbf{T}}_0, \bar{\mathbf{S}}_0 \xleftarrow[\mathcal{D},\mathcal{F}_{shp}]{\text{decode}} \bar{\mathbf{z}}_0$
\State $\mathcal{G}_{id}^{cos} \leftarrow ( 1 - cos(\textbf{V}_{trgt}, \mathcal{C}(\bar{\mathbf{I}}_0 ))  $
\State $\mathcal{G}_{id}^{per} \leftarrow \sum_i \frac{\mathcal{C}^i(\bar{\mathbf{I}}_0) \cdot \mathcal{C}^i(\mathbf{I})} 
    {H_{\mathcal{C}^i} \cdot W_{\mathcal{C}^i} \cdot C_{\mathcal{C}^i} }$
\State $\mathcal{G}_{mse} \leftarrow \| \bar{\mathbf{I}}_0 - {\mathbf{I}}\|_2$
\State $
\mathcal{G}_{lan} \leftarrow \| \mathcal{M}(\bar{\mathbf{I}}_0) - \mathcal{M}({\mathbf{I}})\|_2$
\State $
\mathcal{G}_{vgg} \leftarrow \| \mathcal{V}(\bar{\mathbf{I}}_0) - \mathcal{V}({\mathbf{I}})\|_2 $
\State $\mathcal{G} = \mathcal{G}_{id}^{cos} + \lambda_1 \cdot \mathcal{G}_{id}^{per} + \lambda_2 \cdot \mathcal{G}_{mse} + \lambda_3 \cdot \mathcal{G}_{lan} + \lambda_4 \cdot \mathcal{G}_{vgg} $
\State $\mathbf{z}_{t-1} \backsim \mathcal{N} ( \mathbf{\mu} - s\mathbf{\Sigma} \nabla_{\textbf{z}_t} \mathcal{G}, \mathbf{\Sigma})  $
\EndFor \\
\Return $\mathbf{z}_0$
\end{algorithmic}
\end{algorithm}

Following Algorithm~\ref{alg:ddpm}, we feed the input image $\mathbf{I}$ into $\mathcal{C}$, to extract the latent identity embedding vector $\mathbf{V}_{trgt}$ and the intermediate activation maps.
On top of that, we conduct an alignment step wherein the scene parameters of $\mathbf{I}$ are extracted by using a face detection network~\cite{Schroff_2015_CVPR} and a facial landmark detection network $\mathcal{M}$~\cite{bulat2017far}.
For each reverse diffusion step $t \in \{T,\cdots,1\}$, we firstly predict the injected noise and the corresponding noised variable $\mathbf{z}_t $.
Then, according to the formula in line 5 of Algorithm~\ref{alg:ddpm}, the initial expected latent vector $\bar{\mathbf{z}}_0$ is estimated, followed by the decoding step.
The estimated initial facial texture  $\bar{\mathbf{T}}_0$ and the estimated initial facial shape $\bar{\mathbf{S}}_0$ are computed using the multi-branch facial texture decoder $\mathcal{D}$ and the PCA model $\mathcal{F}_{shp}$, respectively.
After being rendered, the expected facial image $\bar{\mathbf{I}}_0$ is generated. 
We compare the identity embedding vectors between the target image $\mathbf{I}$ and the expected facial image $\bar{\mathbf{I}}_0$ by using the identity cosine distance and identity perceptual loss as defined in \cite{lattas2023fitme}.
Finally, we obtain accurate illumination and facial expression parameters by penalizing the disparity between the per-pixel color intensity and the 3D facial landmarks using 
$\mathcal{G}_{mse} = \parallel \bar{\mathbf{I}}_0 - \mathbf{I} \parallel_2$,
$
\mathcal{G}_{vgg} = \| \mathcal{V}(\bar{\mathbf{I}}_0) - \mathcal{V}({\mathbf{I}})\|_2 $,
and $\mathcal{G}_{lan} = \parallel \mathcal{M}(\mathbf{I}_0) - \mathcal{M}(\mathbf{I}) \parallel_2$.




\section{Controlling the generated identity}
\def \var {0.4}
\begin{figure}[!t]
    \centering
    \includegraphics[width=\var\textwidth]{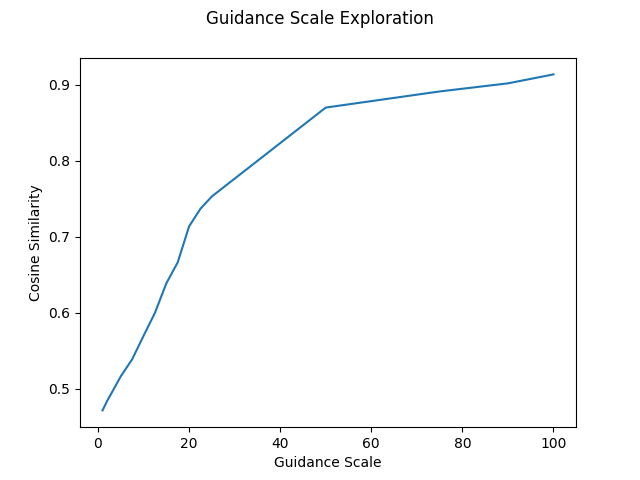}
    \caption{Guidance scale exploration: We randomly pick 10 facial images across the web. We measure the identity similarity between the ground truth image and the generated avatars for different guidance scales.}
    \label{fig:guidance_scale}
\end{figure}

\def \var {0.15}
\begin{figure*}[!t]
\centering
 \begin{subfigure}[b]{\var\textwidth}
      \includegraphics[width=\textwidth]{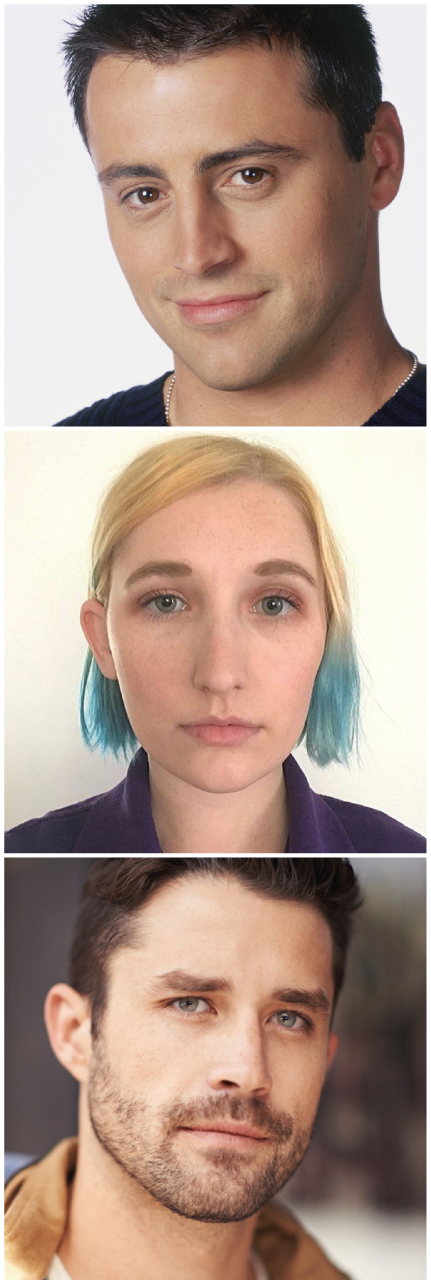}
      \caption{ \centering  Input \\ Facial Images }
      \label{fig:texture-comp-a}
 \end{subfigure}
  \begin{subfigure}[b]{\var\textwidth}
      \includegraphics[width=\linewidth]{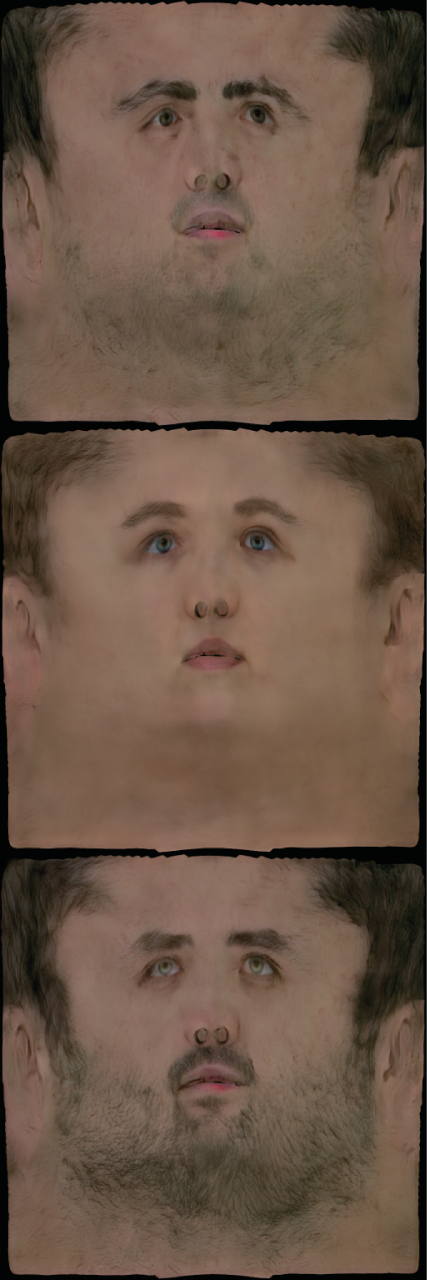}
      \caption{\centering Acquired \\ Diffuse Albedo ($\mathbf{A}_D$)}
      \label{fig:texture-comp-b}
 \end{subfigure}
  \begin{subfigure}[b]{\var\textwidth}
      \includegraphics[width=\linewidth]{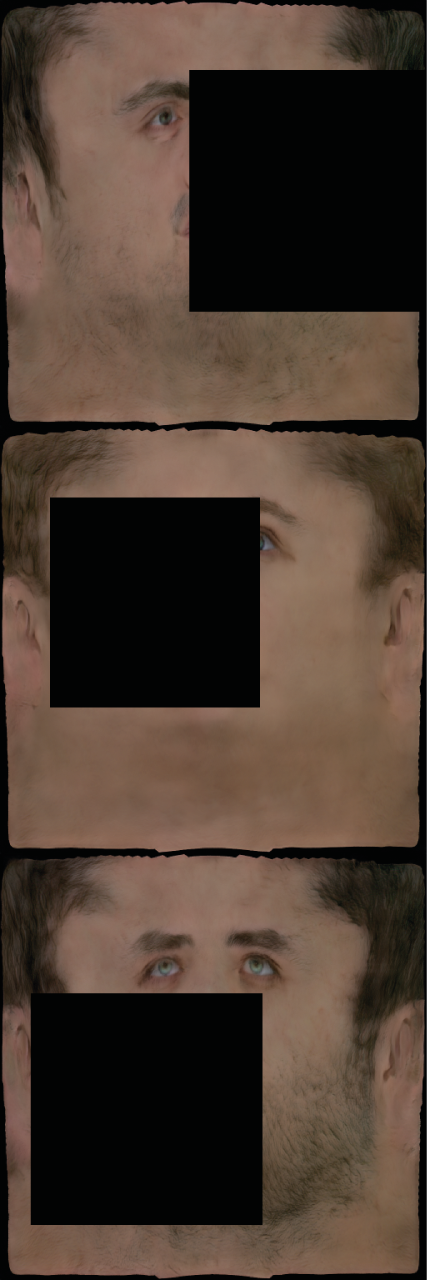}
      \caption{\centering Partial \\ Diffuse Albedo ($\mathbf{A}_D$)}
      \label{fig:texture-comp-c}
 \end{subfigure}
  \begin{subfigure}[b]{\var\textwidth}
      \includegraphics[width=\linewidth]{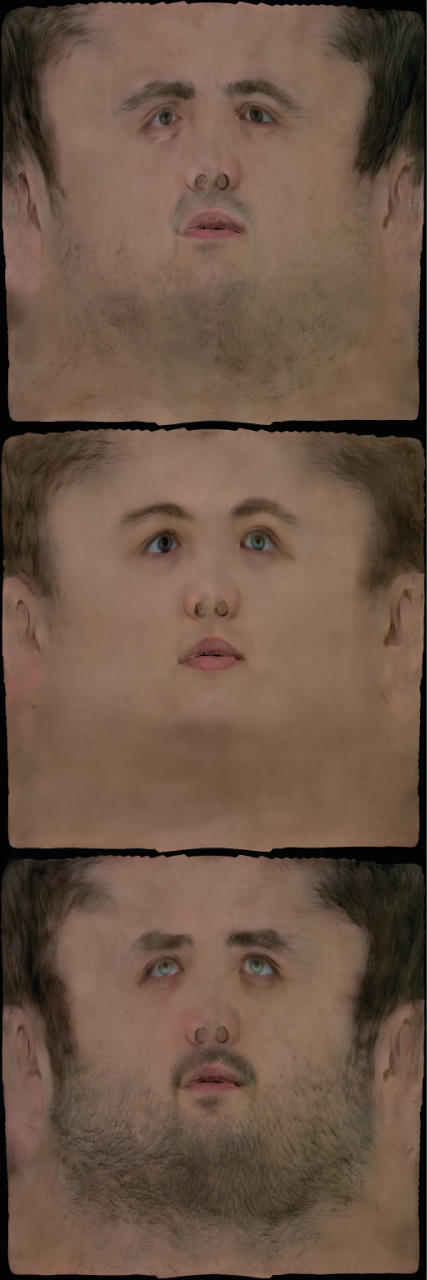}
      \caption{\centering Final \\ Diffuse Albedo ($\mathbf{A}_D$)}
      \label{fig:texture-comp-d}
 \end{subfigure}
  \begin{subfigure}[b]{\var\textwidth}
      \includegraphics[width=\linewidth]{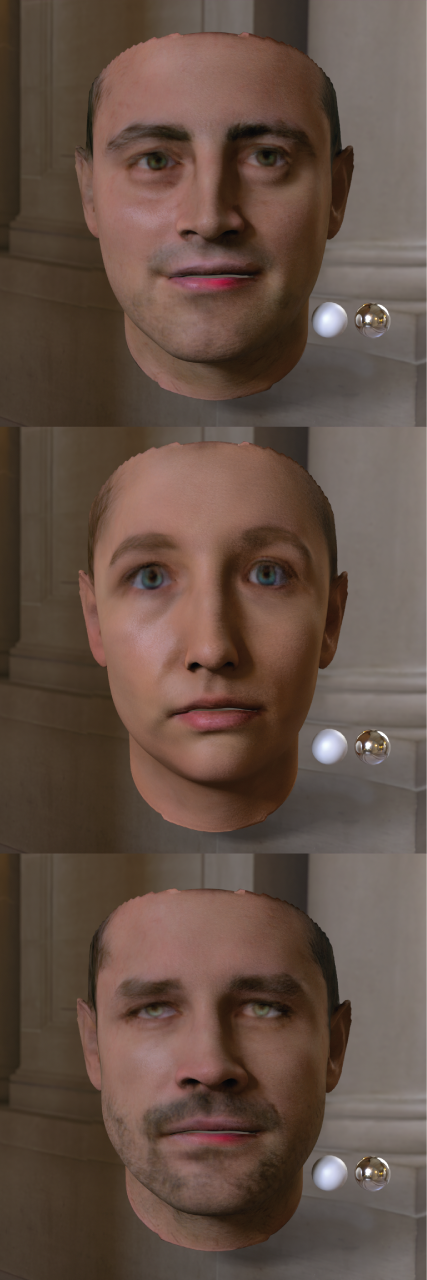}
      \caption{\centering  Rendering\\
      using acquired $\mathbf{A}_D$}
      \label{fig:texture-comp-e}
 \end{subfigure}
  \begin{subfigure}[b]{\var\textwidth}
      \includegraphics[width=\linewidth]{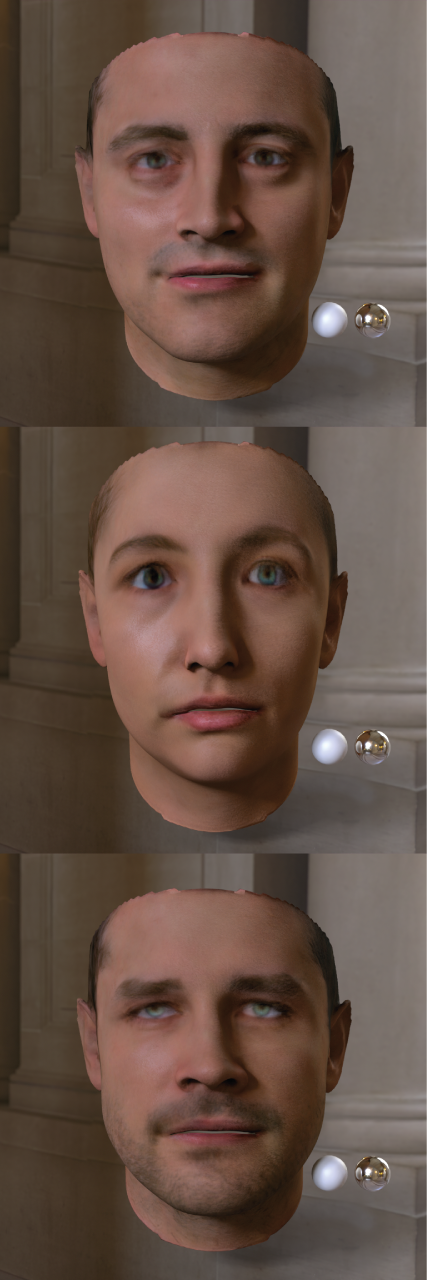}
      \caption{\centering Rendering
      \\using completed $\mathbf{A}_D$}
      \label{fig:texture-comp-f}
 \end{subfigure}
    \caption{Our method can be used for facial texture completion.}
    \label{fig:texture_completion}
\end{figure*}

Choosing the guidance scale is an important factor for the trade-off between the intra-class diversity of the generated samples and the accuracy of the reconstruction.
We conduct an experiment by choosing 10 ``in-the-wild'' images across the web and sample while using different guidance scales for a range of $s = \left[0, 100 \right]$. 
We showcase the results in Fig.~\ref{fig:guidance_scale}.

\section{Partial Texture Completion}
\label{sec:partial}

Inspired by the in-painting approach presented in Relightify~\cite{papantoniou2023relightify}, \ourname{} finds another application in the domain of partial reflectance map completion, illustrated in Fig.~\ref{fig:texture_completion}. 
In certain scenarios, the input reflectance map may be provided partially completed.
Due to the absence of  ground truth facial reflectance maps and with the intention of demonstrating our model's ability to complete partial texture maps, we examine the following scenario:
Given the input images illustrated in Fig.~\ref{fig:texture-comp-a}, we firstly reconstruct the corresponding facial identity (Fig.~\ref{fig:texture-comp-b} and ~\ref{fig:texture-comp-e}).
The resulting diffuse albedo images are treated as pseudo-ground truth and a part of it is randomly masked (Fig.~\ref{fig:texture-comp-c}).
Obtaining completed diffuse albedo maps involves sampling while using only the input identity embedding vector.
The resulting diffuse albedos are showcased in Fig.~\ref{fig:texture-comp-d} whilst the corresponding renderings are shown in Fig~\ref{fig:texture-comp-f}.
By comparing those figures, it is evident that \ourname{} clearly retrieves the masked parts, effectively completing the partially visible reflectance maps.

\def \var {1}
\begin{figure*}[!t]
    \centering
    \begin{subfigure}[b]{\linewidth}
    \includegraphics[width=\textwidth]{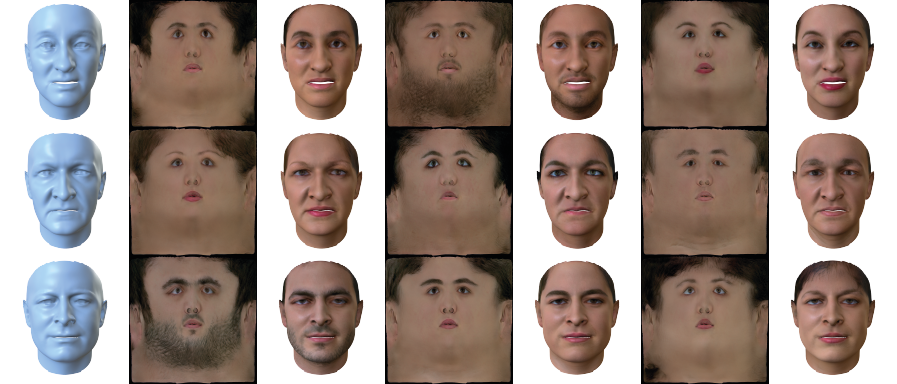}
    \begin{scriptsize}
        \makebox[0.15\linewidth][c]{ Input Shape}\hfill
        \makebox[0.85\linewidth][c]{ Samples}\hfill
    \end{scriptsize}
    
    \caption{Examples of identities generated using the same shape}
    \end{subfigure}
    \begin{subfigure}[b]{\linewidth}
    \includegraphics[width=\textwidth]{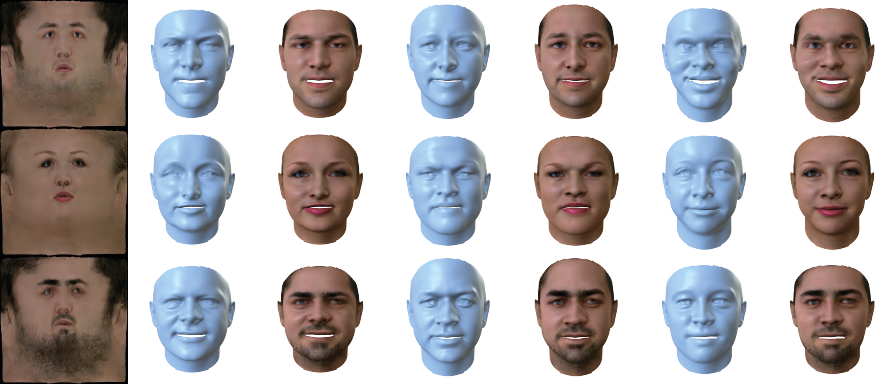}
    \begin{scriptsize}
        \makebox[0.15\linewidth][c]{ Input Texture}\hfill
        \makebox[0.85\linewidth][c]{ Samples}\hfill
    \end{scriptsize}
    \caption{Examples of identities generated having the same texture}
    \end{subfigure}
    \caption{\ourname{} can efficiently disentangle the facial shape and reflectance maps.}
    \label{fig:disentangle}
\end{figure*}

\section{Disentanglement Control}
Another application of the proposed guidance method in Relightify~\cite{papantoniou2023relightify} is used for examining the disentanglement abilities of our proposed approach. More specifically, we consider the following scenarios a) given the facial texture maps as input, \ourname{} generates unconditional facial shapes b) given the input facial shapes as input,  our method generates facial reflectance maps. We present some results in Fig.~\ref{fig:disentangle}.

\section{Additional Results}

\subsection{Shape Reconstruction - REALY benchmark}

\begin{table*}[!t]
    \centering
    \begin{tabular}{|l|c|c|c|c|c|c|}
    \hline
    {} & \multicolumn{3}{c}{@Nose} & \multicolumn{3}{|c|}{@Mouth}\\ 
    Method & avg & med & std & avg & med & std \\
    \hline
    HiFace-f~\cite{Chai_2023_ICCV} & \textbf{1.036} & \textbf{0.992} & \textbf{0.280} & 1.450 & 1.388 & 0.413 \\
    \hline
    HiFace-c~\cite{Chai_2023_ICCV} & 1.054 & 1.021 & 0.317 & 1.461 & 1.381 & 0.430 \\
    \hline
    HRN~\cite{lei2023hierarchical}& 1.722 & 1.685 & 0.330 & \textbf{1.357} & \textbf{1.226} & 0.523  \\
    \hline 
    Deep3D~\cite{deng2019accurate}& 1.719 & 1.683 & 0.354 & 1.368 & 1.301 & 0.439\\
    \hline
    AlbGAN~\cite{rai2023towards} & 1.656 & 1.636 & 0.374 & 2.087 & 1.927 & 0.839\\
    \hline
    MGCNet~\cite{shang2020self} & 1.771 & 1.741 & 0.380 & 1.417 & 1.355 & 0.409\\
    \hline
    GANFit~\cite{Gecer_2019_CVPR} & 1.928 & 1.881 & 0.490 & 1.812 & 1.769 &0.544\\
    \hline
    FitMe~\cite{lattas2023fitme} & 1.833 & 1.796 & 0.434 & 1.752 & 1.629 &0.539\\
    \hline
    PSL~\cite{https://doi.org/10.1111/cgf.14945} & 1.708 & 1.688 & 0.349 & 1.708 & 1.777 & 0.563\\
    \hline 
    DECA-c~\cite{10.1145/3450626.3459936} & 1.697 & 1.654 & 0.355 & 2.516 & 2.465 & 0.839\\
    \hline
    CEST~\cite{Wen2021SelfSupervised3F} & 2.779 & 2.717 & 0.835 & 1.448 & 1.438 & \textbf{0.406}\\
    \hline
    EMOCA-c\cite{EMOCA:CVPR:2021} & 1.868 & 1.821 & 0.387 & 2.679 & 2.419 & 1.112\\
    \hline
    MICA~\cite{MICA:ECCV2022} & 1.585 & 1.542 & 0.325 & 3.478 & 3.439 & 1.204\\
    \hline
    DECA-f\cite{10.1145/3450626.3459936} & 2.138 & 2.137 & 0.461 & 2.802 & 2.699 & 0.868\\
    \hline
    EMOCA-f~\cite{EMOCA:CVPR:2021} &  2.532 & 2.563 & 0.539 & 2.929 & 2.676 & 1.106\\
    \hline
    \ourname{}(Ours) & 1.821 & 1.770 & 0.438 & 1.751 & 1.611 & 0.523\\
    \hline
      \end{tabular}
    \caption{Results in the REALY benchmark~\cite{REALY}}
    \label{tab:really-1}
\end{table*}

\begin{table*}[!t]
    \centering
    \begin{tabular}{|l|c|c|c|c|c|c|c|}
    \hline
    {} & \multicolumn{3}{c}{@Forehead} & \multicolumn{3}{|c|}{@Cheek} & All\\
    Method & avg & med & std & avg & med & std & avg \\
    \hline
    HiFace-f~\cite{Chai_2023_ICCV} & \textbf{1.324} & \textbf{1.296} & \textbf{0.334} & 1.291 & 1.240 & 0.362 & \textbf{1.275}\\
    \hline
    HiFace-c~\cite{Chai_2023_ICCV} & 1.331 & 1.307 & 0.347 & 1.342 & 1.304 & 0.384 & 1.297\\
    \hline
    HRN~\cite{lei2023hierarchical}& 1.995 & 1.990 & 0.476 & \textbf{1.072} & 1.063 & 0.333 & 1.537 \\
    \hline 
    Deep3D~\cite{deng2019accurate}& 2.015 & 2.007 & 0.449 & 1.528 & 1.442 & 0.501 & 1.657 \\
    \hline
    AlbGAN~\cite{rai2023towards} & 2.102 & 2.035 & 0.512 & 1.141 & 1.103 & \textbf{0.303} & 1.746\\
    \hline
    MGCNet~\cite{shang2020self} & 2.268 & 2.215 & 0.503 & 1.639 & 1.494 & 0.650 &1.774 \\
    \hline
    GANFit~\cite{Gecer_2019_CVPR} & 2.402 & 3.339 & 0.545 & 1.329 & 1.234 & 0.504 & 1.868 \\
    \hline
    FitMe~\cite{lattas2023fitme} & 2.494 & 2.385 & 0.605 & 1.414 & 1.315 & 0.526 & 1.873 \\
    \hline
    PSL~\cite{https://doi.org/10.1111/cgf.14945} & 2.350 & 2.343 & 0.551 & 1.593 & 1.482 & 0.540 &1.882\\
    \hline 
    DECA-c~\cite{10.1145/3450626.3459936} & 2.394 & 2.256 & 0.576 & 1.479 & 1.400 & 0.535 & 2.010\\
    \hline
    CEST~\cite{Wen2021SelfSupervised3F} & 2.384 & 2.302 & 0.578 & 1.456 & 1.321 & 0.485 & 2.017 \\
    \hline
    EMOCA-c\cite{EMOCA:CVPR:2021} & 2.426 & 2.383 & 0.641 & 1.438 & 1.294 & 0.501 & 2.103\\
    \hline
    MICA~\cite{MICA:ECCV2022} & 2.374 & 2.251 & 0.683 & 1.099 & \textbf{1.003} & 0.324 & 2.134 \\
    \hline
    DECA-f\cite{10.1145/3450626.3459936} & 2.457 & 2.341 & 0.559 & 1.443 & 1.353 & 0.498 & 2.210\\
    \hline
    EMOCA-f~\cite{EMOCA:CVPR:2021} & 2.595 & 2.505 & 0.631 & 1.495 & 1.360 & 0.469 & 2.388 \\
    \hline    
    \ourname{}(Ours) & 2.472 & 2.322 & 0.581 & 1.404 & 1.287 & 0.525 & 1.862 \\
    \hline
      \end{tabular}
    \caption{Results in the REALY benchmark~\cite{REALY}}
    \label{tab:really-2}
\end{table*}

We evaluate our method's shape reconstruction with state-of-the-art methods using REALY~\cite{REALY}, a widely used public benchmark.
It contains 100 high-quality face shapes from different ethnic and age backgrounds, based on the LYHM dataset~\cite{LYHM}. 
Contrary to previous face geometry reconstruction challenges~\cite{RingNet:CVPR:2019}, the REALY benchmark computes geometric errors separately for each region of the human face while using the $ l_2$ distance between the ground truth and the predicted meshes.
The results of this benchmark are showcased in Tab.~\ref{tab:really-1} and \ref{tab:really-2}. Our method ranks 7th on the average reconstruction error and gets surpassed only by models
that either focus solely on generating facial shape (HiFace~\cite{Chai_2023_ICCV}). or produce a single albedo map with baked-in illumination
(HRN~\cite{lei2023hierarchical}, Deep3D~\cite{deng2019accurate}, AlbGAN~\cite{rai2023towards}, MGCNet~\cite{shang2020self}), which restricts the
resulting avatars from being relightable. 

Our model's performance can be explained by the fact that our approach is trained using synthetic data obtained from a fitting methodology~\cite{lattas2023fitme} and generates both albedo and normals UV maps.
The utilization of synthetic data imposes inherent limitations on our method's ability to accurately retrieve facial shapes. This limitation stems from the constraints imposed by FitMe on shape retrieval performance. We eventually beat FitMe's performance, as well as similar methods (e.g. GANFit), showing that our results are bounded by the training data, and could improved given a real captured dataset.
Additionally, as highlighted by the authors of FitMe, the concurrent reconstruction of both facial shape and texture normals introduces further constraints on the approach's shape reconstruction performance.
Specifically, a portion of the shape information is occasionally encoded in the normals domain rather than in the actual facial shape domain.

\subsection{Comparison between a single model with separate models}
In this section we analyze the effectiveness of a unified single-model architecture compared to a multi-model approach.
Using a randomly selected subset of 50 identities from the REALY benchmark~\cite{REALY}, we fit our model following two distinct strategies: a) independently sampling for facial shape and reflectance maps, and b) employing our proposed methodology.
We then evaluate their identity similarity scores and the facial shape reconstruction performance by comparing the generated facial avatars against the ground truth using the evaluation pipeline provided by the REALY benchmark~\cite{REALY} . 
The results are presented in Tab.~\ref{tab:single_model}, and demonstrate that the unified single-model approach achieves superior performance in both metrics.

\begin{table}[h]
\centering
\begin{scriptsize}
\setlength{\tabcolsep}{5pt}
\centering
\begin{tabular}{lccc}
\toprule
\textbf{Method}  & Facial Shape $\downarrow$  & ID similarity $\downarrow$ \\ 
\midrule
\textbf{Separate Models} & 1.805 & 0.873 \\
\midrule
\textbf{\ourname{}(Ours)} & \textbf{1.764} & \textbf{0.911} \\
\bottomrule
\vspace{0.12cm}
\end{tabular}

\captionof{table} {Comparison between the single model approach (\ourname{}) and using 2 separate models for facial reconstruction}
 \label{tab:single_model}
\end{scriptsize}
\end{table}

{\small
\bibliographystyle{ieee_fullname}
\bibliography{egbib}
}